\documentclass[journal=prl, 11pt, notitlepage, reprint, amsmath,nonatbib]{revtex4-2}

\usepackage{graphicx}
\usepackage{amsmath}
\usepackage[normalem]{ulem}
\usepackage{hyperref}
\usepackage{subcaption}
\usepackage[utf8]{inputenc} % allow utf-8 input
\renewcommand{\selectlanguage}[1]{}
\newcommand{\figspace}{\vspace{-6pt}}
\newcommand{\bigO}{\mathcal{O}}
\newcommand{\diff}[1]{#1}

\usepackage[T1]{fontenc}    % use 8-bit T1 fonts
\usepackage{hyperref}       % hyperlinks
\usepackage{url}            % simple URL typesetting
\usepackage{booktabs}       % professional-quality tables
\usepackage{amsfonts}       % blackboard math symbols
\usepackage{nicefrac}       % compact symbols for 1/2, etc.
\usepackage{microtype}      % microtypography
\usepackage{xcolor}         % colors
\usepackage[labelfont=bf]{caption}

\begin{document}

\title{Interpretable Visualizations of Data Spaces for Classification Problems}

\author{Christian Jorgensen}
\affiliation{Department of Chemical and Biological Engineering, University of Wisconsin - Madison, Madison, WI, USA}
\author{Arthur Y. Lin}
\affiliation{Department of Chemical and Biological Engineering, University of Wisconsin - Madison, Madison, WI, USA}
\author{Rhushil Vasavada}
\affiliation{Department of Computer Sciences, University of Wisconsin - Madison, Madison, WI, USA}
\author{Rose K. Cersonsky}
\email{rose.cersonsky@wisc.edu}
\affiliation{Department of Chemical and Biological Engineering, University of Wisconsin - Madison, Madison, WI, USA}
\affiliation{Department of Materials Science and Engineering, University of Wisconsin - Madison, Madison, WI, USA}
\affiliation{Data Science Institute, University of Wisconsin - Madison, Madison, WI, USA}

\begin{abstract}
How do classification models ``see'' our data? Based on their success in delineating behaviors, there must be some lens through which it is easy to see the boundary between classes; however, our current set of visualization techniques makes this prospect difficult. In this work, we propose a hybrid supervised-unsupervised technique distinctly suited to visualizing the decision boundaries determined by classification problems. This method provides a human-interpretable map that can be analyzed qualitatively and quantitatively, which we demonstrate through several established examples from literature. While we discuss this method in the context of chemistry-driven problems, its application can be generalized across subfields for ``unboxing'' the operations of machine-learning classification models.
\end{abstract}

\maketitle

\section{Introduction}

Over the past decades, machine learning (ML) has transformed computational chemistry, for example accelerating materials discovery \cite{liu_materials_2017}, improving the quality of molecular simulations \cite{bartok_gaussian_2010, behler_generalized_2007}, and enabling reliable predictions of materials properties \cite{pilania_accelerating_2013}. However, many traditional ML and deep learning methods function as ``black boxes,'' providing little insight into the underlying physics and chemistry of materials and materials processes. In chemical research, the true value of an ML model often goes beyond its ability to make accurate predictions; it also lies in its ability to reveal how the set of input parameters, often carefully chosen to capture the underlying chemistry, relate to the property of interest. Addressing this need for interpretability has driven the development of approaches specifically tailored to uncover structure-property relationships, a pursuit central to the field of Quantitative Structure-Property Relationship (QSPR) modeling. 

Many approach QSPR through human-interpretable visualization via unsupervised machine-learning methods. Principal Component Analysis (PCA) is the most standard method for mapping tasks \cite{pearson_liii_1901}. PCA involves identifying a linear (or nonlinear, in the case of \textit{kernel} PCA) combination of the features that maximizes the variance (i.e., information) retained. Beyond PCA, there exist a slue of other mapping techniques commonly used to qualitatively understand data spaces -- t-distributed Stochastic Neighbor Embedding (t-SNE) \cite{maaten_visualizing_2008, policar_opentsne_2024}, Uniform Manifold Approximation and Projection (UMAP) \cite{mcinnes_umap_2020}, and Isometric Mapping (ISOMAP) \cite{tenenbaum_global_2000} focus on the preservation of the local, global, and geometric data structure, respectively, and Locally Linear Embedding (LLE) \cite{roweis_nonlinear_2000} and multi-dimensional scaling (MDS) \cite{kruskal_multidimensional_1964, kruskal_nonmetric_1964} preserve distances between neighboring data points. While all of these methods excel at creating informative visualizations and identifying patterns in the structure of the data, the ``P'' (property) of ``QSPR'' is only spuriously accounted for, where any relationship between the visualized dimensions and the property of interest is purely coincidental.

So, how do we determine which chemical or materials characteristics most influence the decisions made by our model?
When these characteristics are explicitly included in our input vectors, feature importance algorithms are indispensable for such tasks. Tree-based methods, such as random forests, are most commonly associated with feature importance analysis, as they provide built-in ``feature importance'' measures, which are typically computed as the total reduction of either the Gini impurity (for classification) or the squared error (for regression) of each of the features \cite{breiman_random_2001}. SHAP (SHapley Additive exPlanations), another popular post-training analysis scheme, draws from game theory and involves assigning each feature a value (i.e., SHAP value) representing its contribution to a model's prediction by considering all possible combinations of feature subsets \cite{lundberg_unified_2017}. Alternatively, many have measured the importance of different features by measuring the performance of feature subsets through iteratively adding or removing features \cite{jennrich_application_1968}. These techniques can excel because they can identify the most relevant descriptors from a large feature set. Still, they may not be particularly informative when the features are heavily correlated or correspond to abstract concepts.

Thus, while methods from unsupervised learning and supervised learning address aspects of QSPR, none can fully do so on their own. Consequently, this problem has motivated the extension of existing unsupervised learning algorithms to incorporate target information into their mapping strategies, with two beneficiaries of this direction having been UMAP and PCA. With UMAP, supervision can be added by intersecting the unsupervised UMAP graph constructed on the input features with a graph constructed on the targets. However, recent works have validated that because unsupervised UMAP is already designed for resolving clusters, adding supervision does not significantly alter performance  \cite{zhai_supervised_2022}. 
Supervised PCA, on the other hand, involves modifying the principal components to instead identify directions in which the dependence between the input and output data is maximized \cite{barshan_supervised_2011, ghojogh_unsupervised_2019}. More specifically, supervised PCA learns a linear (or kernel, in kernel supervised PCA) projection that maximizes an empirical approximation of the Hilbert-Schmidt independence criterion \cite{gretton_measuring_2005}, which itself is a function of the cross-covariance between the input and output data. While this formulation ensures that classes would be well-separated, its major shortcoming is that it does not explicitly preserve the variance of the input data. This can result in cases in which directions in the data that are correlated with the target due to collinearity are overrepresented in the map, despite the fact that these directions are likely noise.

Concurrently, there has been an emergence of ``visualization as interpretation'' techniques that aim to provide a solution. Briefly, the vast majority of these methods aim to compute invertible mappings that represent the partitioning and clustering of data in higher-dimensional spaces \cite{wang_quantitative_2023,rodrigues_constructing_2019,oliveira_sdbm_2022,blumberg_multiinv_2025}. For example, DeepView, the most notable of these methods, combines UMAP and Fisher distance for projection and then samples the resultant latent space on a grid to assess data boundaries \cite{schulz_deepview_2020}. However, computational cost and model complexity have remained large challenges for many of these methods, particularly those reliant on multi-layer architectures or grid-based assessments. And, methods based on high-nonlinear operations or architectures can result in topologically-bound latent spaces, where arbitrary points or pixels in a grid may not directly correspond to physically realizable data points \cite{wang_fundamental_2024}.

We address these challenges by exploiting a family of methods, known as \textit{hybrid} learning, that draws elements from both supervised and unsupervised learning. Some of these methods include principal components regression \cite{jolliffe_note_1982}, partial least squares regression \cite{wold_pls-regression_2001}, cluster-wise regression \cite{spath_algorithm_1979}, continuum regression \cite{stone_continuum_1990}, and principal covariates regression (PCovR) \cite{de_jong_principal_1992}. Of these methods, PCovR is particularly valuable for QSPR because it is directly formulated as a multi-objective problem in which the regression and dimensionality reduction tasks are optimized simultaneously. PCovR has broad applications within the fields of climate science \cite{fischer_regularized_2014}, economic forecasting \cite{heij_forecast_2007}, psychology \cite{vervloet_pcovr_2015}, bioinformatics \cite{cersonsky_placental_2023, van_deun_obtaining_2018, taylor_nutrition_2019}, and molecular crystallization \cite{cersonsky_data-driven_2023}. Similar to PCA, PCovR can also be used to incorporate nonlinear relationships through similarity kernels \cite{helfrecht_structure-property_2020}, and can be used with regularization to create sparse solutions \cite{fischer_regularized_2014, van_deun_obtaining_2018}. In addition, PCovR has been used to modify other techniques, such as clusterwise regression \cite{wilderjans_principal_2017}, farthest point sampling, and CUR decomposition \cite{cersonsky_improving_2021}.

Despite its versatility and wide applicability, PCovR, along with most other hybrid learning methods, is primarily used for regression tasks, where the output is continuous. Yet, many important materials problems require us to determine relationships between design principles and \emph{discrete} properties, as are common in classification problems. For classification, however, the lack of a method that integrates dimensionality reduction and classification performance makes QSPR difficult to perform. We propose a new method for QSPR, which we call principal covariates classification (PCovC), that involves using PCovR as an intermediary for predicting class likelihoods. 

\section{Theory}
\label{sec:theory}

We provide a concise description of the underlying mathematics for principal covariates classification (PCovC). We direct interested readers to \cite{helfrecht_structure-property_2020} for a more thorough discussion, including that of data scaling and incorporating nonlinearity. Included in App.~B are discussions of other QSPR techniques, namely SHAP analysis \cite{lundberg_unified_2017}, linear discriminant analysis \cite{hastie_elements_2009}, and saliency maps \cite{simonyan_deep_2014}.

\subsection{Notation}
\begin{figure*}[ht!]
    \centering
    \includegraphics[width=\linewidth]{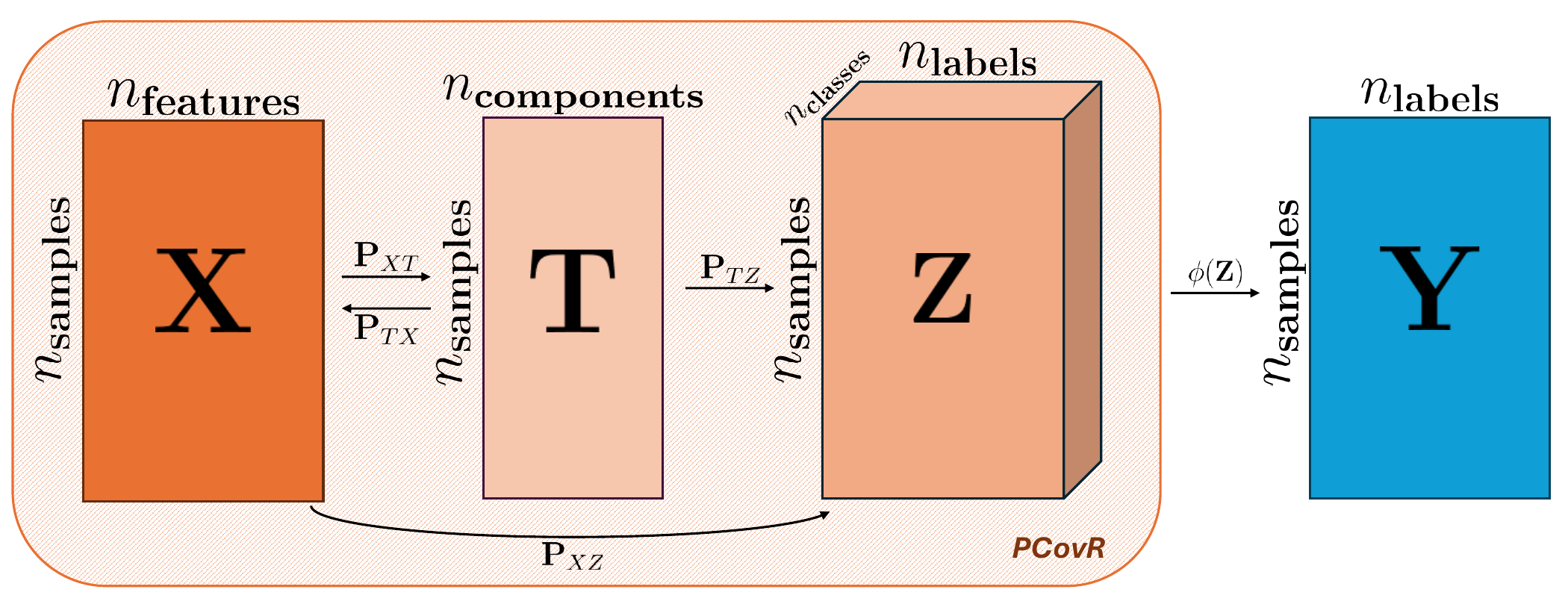}
    \caption{\textbf{Relationship between variables and projectors in principal covariates classification.} $\mathbf{Z}$ is the evidence matrix, which is a quantification of class likelihoods, and it can be a tensor in the case of a multilabel, multiclass classification problem.}
    \label{fig:pcovc_relationships}\figspace
\end{figure*}

We assume that the data is organized into two objects: a ``feature matrix'', $\mathbf{X}$, that contains a set of input variables for a set of observations, and a ``target matrix'', $\mathbf{Y}$, that contains the classes for each observation. The rows of the feature matrix $\mathbf{X}$ correspond to the individual samples and the columns correspond to the features, leading to shape \(n_{\text{samples}}\times n_{\text{features}}\). The target matrix $\mathbf{Y}$ has shape \(n_\text{{samples}}\times n_\text{{labels}}\) and contains the output labels for each observation, each of which can take an integer value between $0$ and $n_{\text{classes}}-1$. We denote the latent-space projection of the input data as $\mathbf{T}$, which has shape $n_{\text{samples}} \times n_{\text{components}}$. Finally, we denote $\mathbf{Z}$ as the \textit{evidence} tensor, which contains confidence scores for each class. $\mathbf{Z}$ has shape of $n_{\text{samples}} \times n_{\text{classes}} \times n_{\text{labels}}$. We assume that $\mathbf{P_{XZ}}$ is the set of weights learned by a trained classification model, where, depending on the architecture of this model, $\mathbf{Z}$ is used to determine the class of the data points. 
In this context, we wish to learn linear mappings between $\mathbf{X}$, $\mathbf{T}$, and $\mathbf{Z}$, which we can define using projection matrices. In this work, we define $\mathbf{P}_{AB}$ as the projector from space $\mathbf{A}$ to space $\mathbf{B}$. Therefore, we can write $\mathbf{T} \equiv \mathbf{XP}_{XT}$ and $\mathbf{Z}=\mathbf{TP}_{TZ}=\mathbf{XP}_{XT}\mathbf{P}_{TZ}.$
The relationship between these objects is shown in Fig.~\ref{fig:pcovc_relationships}.

\subsection{Linear Classification}
The goal of a classification model is to assign an unseen sample to one of $n_{\text{classes}}$ discrete classes. Thus, we aim to draw ``decision boundaries'' that divide the input space into $n_{\text{classes}}$ regions, with each region corresponding to a certain class. In a linear classification problem, these boundaries are lines or hyperplanes in a higher-dimensional space. Deterministic approaches, such as support vector machines, involve finding a discriminant function and use an activation function to map this function to a specific class label. Probabilistic approaches, such as logistic regression, estimate the probability distributions of the input and output and use these probabilities to make classifications.
More generally, linear classifiers make predictions by learning a linear projection of the input data $\mathbf{Z}=\mathbf{XP}_{XZ},$
and then using some function $\phi$, typically called an activation function, to map the continuous $\mathbf{Z}$ to the set of discrete class labels $\mathbf{Y}\in{0, 1, 2,  ... n_{\text{classes}}-1}$, such that $\mathbf{Y}=\phi(\mathbf{Z}).$

Linear classifiers are not typically used for visualization, with one strong exception: linear discriminant analysis, or LDA. LDA determines a linear projector $\mathbf{P}_{XT}$ of a given dataset that optimizes the ratio of inter-cluster scatter $\mathbf{S}_B\equiv \sum_c N_c \left( \langle \mathbf{X}_c\rangle - \langle \mathbf{X} \rangle \right) \left( \langle \mathbf{X}_c\rangle - \langle \mathbf{X} \rangle\right)^T$ and intra-cluster scatter $\mathbf{S}_W\equiv \sum_c \sum_{i \in c} \left( \mathbf{x}_i - \langle \mathbf{X}_c\rangle \right) \left( \mathbf{x}_i - \langle \mathbf{X} \rangle_c \right)^T$, where $N_c$ is the number of points in cluster $c$, $\langle\mathbf{X}_c\rangle$ is the column-wise average of features of points belonging to cluster $c$, and $x_i$ is the feature vector for point $i$ belonging to cluster $c$. The optimization function $\mathcal{J}(\mathbf{P}_{XT})$ is then written as:\vspace{-6pt}

\[
\mathcal{J}(\mathbf{P}_{XT}) = \frac{ \left| \mathbf{P}_{XT}^T \mathbf{S}_B \mathbf{P}_{XT} \right| }{ \left| \mathbf{P}_{XT}^T \mathbf{S}_W \mathbf{P}_{XT} \right| }
\]

\vspace{-3pt}This will yield an $n_\text{classes}-1$ projection of the data that can be analyzed for quantitative structure-property relations. We have included comparisons with LDA within the text to demonstrate the abilities and shortcomings of this method. Further discussion of LDA can be found in App.~B.2.

\subsection{Principal Covariates Regression}

PCovR \cite{de_jong_principal_1992, helfrecht_structure-property_2020} is a mapping technique to determine a low-dimensional $\mathbf{T}\equiv \mathbf{X P}_{XT}$ that minimizes

\begin{equation} \label{eq:pcovr_objective}
\begin{aligned}
\ell = \alpha\underbrace{||\mathbf{X-XP}_{XT}\mathbf{P}_{XT}^T||^2}_{\text{loss in reconstructing X}}+(1-\alpha)\underbrace{||\mathbf{Y}-\mathbf{XP}_{XT}\mathbf{P}_{TY}||^2}_{\text{loss in predicting Y}},
\end{aligned}
\end{equation}

where a mixing parameter $\alpha$ is used to determine the relative weight of data reconstruction versus property regression.
For linear and kernel ridge regression, our projector from $\mathbf{T} $ ($\mathbf{P}_{TY}$) is an analytically solvable function of $\mathbf{X}$ and $\mathbf{Y}$, and thus Eq.~\eqref{eq:pcovr_objective} is analytically solvable by performing an eigendecomposition of a modified Gram matrix $\mathbf{\tilde{K}} = \alpha \mathbf{XX}^T + (1 - \alpha)\mathbf{\hat{Y}\hat{Y}}^T,$
or, equivalently, a modified covariance matrix $\mathbf{\tilde{C}} = \alpha \mathbf{X}^T\mathbf{X} + (1 - \alpha) \left( \left( \mathbf{X}^T \mathbf{X} \right)^{-\frac{1}{2}} \mathbf{X}^T \mathbf{\hat{Y} \hat{Y}}^T \mathbf{X} \left( \mathbf{X}^T \mathbf{X} \right)^{-\frac{1}{2}} \right)
\label{eq:pcovr_covariance}, $
where $\mathbf{\hat{Y}} = \mathbf{XP}_{XY}$ is the linear regression approximation of $\mathbf{Y}$, used to prevent overfitting. This procedure is easily adapted for methods based on data similarity, i.e., kernel models, as detailed in \cite{helfrecht_structure-property_2020}.

\begin{figure*}[t]
\centering
    \includegraphics[trim = {0cm, 0cm, 0cm, 0cm}, clip, width=\linewidth]{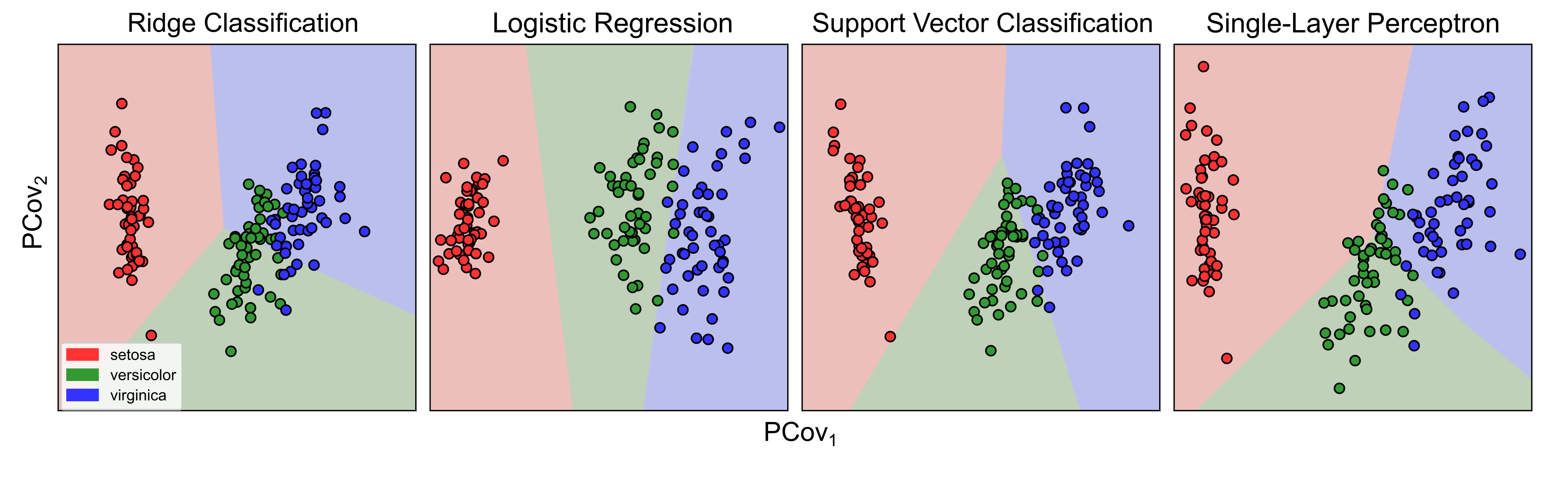}
    \caption{\textbf{PCovC maps will change based on the underlying classifier.} PCovC ($\alpha=0.5$) maps for the Iris dataset \cite{fisher_use_1936, anderson_species_1936} with ridge classification, logistic regression, support vector classification, and a single-layer perceptron. Color corresponds to the class (red, green, and blue for setosa, versicolor, and virginica flower classes) and the background shows the estimated decision boundaries.}
 \label{fig:pcovc_classifers}\figspace
\end{figure*}
\subsection{Principal Covariates Classification}
The PCovR loss function, Eq.~\eqref{eq:pcovr_objective}, should not be directly modified to be used in classification tasks -- replacing the regression loss with a classification loss often results in a non-analytical function (when the underlying classification algorithm incorporates activation functions into their loss optimization) or is scientifically non-productive, as many classification tasks, particularly multiclass tasks, use numerical values to denote class, where these numerical values are ostensibly arbitrary.
We choose to adapt PCovR for classification tasks by leveraging the evidence as our approximation of $\mathbf{Y}$. This yields a new objective function
\begin{equation} \label{eq:pcovc_objective}
\begin{aligned}
\ell = \alpha\underbrace{||\mathbf{X-XP}_{XT}\mathbf{P}_{XT}^T||^2}_{\text{loss in reconstructing X}}+(1-\alpha)\underbrace{||\mathbf{Z}-\mathbf{XP}_{XT}\mathbf{P}_{TZ}||^2}_{\text{loss in predicting Z}},
\end{aligned}
\end{equation}

and a new modified Gram matrix (or modified covariance matrix)

\begin{equation} 
\mathbf{\tilde{K}} = \mathbf{\alpha XX}^T + (1 - \alpha) \mathbf{ZZ}^T
\label{eq:pcovc_gram}
\end{equation}

\noindent whose eigendecomposition can be used to obtain a relevant low-dimensional map. Similar to PCovR, this procedure can be adapted for nonlinear kernel classification models, where instead $\mathbf{Z}\equiv \mathbf{KP}_{KZ}$.

Thus, an important component of this procedure is the selection of the expectation matrix $\mathbf{Z}$. 
As different classification algorithms operate to optimize different loss terms, they will yield different values of $\mathbf{Z}$; this is visualized in Fig.~\ref{fig:pcovc_classifers}, where we demonstrate the new latent space maps determined for PCovC at $\alpha=0.5$ for the archetypal Iris dataset \cite{fisher_use_1936, anderson_species_1936} and several popular linear classification algorithms. These maps are well-founded, as, by construction, there exists a strong relationship between $\mathbf{X}$ and $\mathbf{Z}$. This framework indeed is strongest where there exists a clear relationship between $\mathbf{X}$ and $\mathbf{Z}$ (or, equivalently, $\mathbf{K}$ and $\mathbf{Z}$). Mathematically, this allows the easy identification of covariances between variables, and thus the loss written in Eq.~\ref{eq:pcovr_objective} to vary smoothly with $\alpha$.

However, this does not impose a necessary requirement that $\mathbf{Z}$ comes from a classification model constructed by $\mathbf{Y} = \phi (\mathbf{Z}),$ simply that $\mathbf{Z}$ reflects the class likelihoods and is normalized such that its variance is of similar magnitude to $\mathbf{X}$ (having $\mathbf{X}$ and $\mathbf{Z}$ of non-standardized variance leads to asymptotic behavior of relevant loss functions with respect to $\alpha$, as discussed in \citet{helfrecht_structure-property_2020}). Also, $\mathbf{Z}$ should be determined in such a way as to prevent over-fitting, as is typically done by regularizing losses in classification models.

Choosing $\alpha$ depends on the intention behind the mapping procedure (how much one cares about the reconstruction of $\mathbf{X}$ versus the clear visualization of the decision boundary), but can be done programmatically based on the mixed loss (Eq.~\ref{eq:pcovc_objective}). When the intention is to understand the behavior of the classifier, it is important to note that for binary classification with $\mathbf{Z}=\mathbf{XP}_{XZ}$ or $\mathbf{Z}=\mathbf{KP}_{KZ}$, $\mathbf{Z}$ should be fully reproducible from $\mathbf{T}$ as $\alpha\to0$. This makes low (but nonzero) $\alpha$ values, such as $0.05$, more attractive, as the classifier's learned decision boundary can be well-captured in this regime while also learning an orthogonal dimension for analysis. With multiclass problems, we require that $\mathbf{T}$ has at least $n_\text{classes}$ dimensions to fully reconstruct $\mathbf{Z}$, thus the same recommendations are not directly applicable.

With perfectly balanced datasets wherein the $\mathbf{X}$ and $\mathbf{Z}$ matrices are standardized to unit variance, $\alpha$ can typically be taken as 0.5, as this weights the reconstruction of the matrices equally. When the matrices are of different variances or the classes are imbalanced, $\alpha$ should be determined by a minimization of PCovC loss, ideally through cross-validation. We have followed this procedure in all subsequent case studies, and have included corresponding cross-validation results in the Supplementary Information. For a more thorough discussion of the impact of $\alpha$, we direct the interested reader to \citet{vervloet_model_2016}.

\section{Results and Discussion}
In the following case studies, we aim to show both the utility and flexibility of this approach across many different topic areas, data representations, and classification architectures. While we aim to demonstrate the extraction of actionable and insightful principles from each of these studies, we encourage the readers to refer to the original articles for more thorough scientific discussion.

\subsection{What makes a molecule toxic? Exposing decision boundaries in toxicity classifications}
\label{sec:tox}

In this case study, we use PCovC to reveal a latent space that discriminates chemicals based on their neurotoxicity. More specifically, we leverage the dataset and classification procedures discussed in \cite{wu_trade-off_2021}, which itself contains a processed subset of 68 bioassay endpoints for 7,660 molecules from the Tox21 dataset \cite{richard_tox21_2021}. We use a subset of this dataset, focusing only on the three assays that measure the potential for compounds to disrupt the acetylcholinesterase pathway. Since acetylcholinesterase is responsible for terminating synaptic transmission in the brain, these assays are considered as good indicators of neurotoxicity \cite{li_profiling_2021}. Within the dataset, these assays are labeled ``tox21-ache-p1 (-1)'', ``tox21-ache-p3 (-1)'', and ``tox21-ache-p5 (-1)''. We will refer to molecules / data points considered ``active antagonists'' for a given assay as ``toxic'' and ``inactive'' as ``non-toxic.'' Following procedure from \cite{wu_trade-off_2021}, we generate the RDKit fingerprint \cite{landrum_rdkitrdkit_2025}, which encodes molecular topology by mapping atom connectivity and local structure into a compact, sparse binary representation. We then use PCovC to create a map that delineates chemicals by the ``tox21-ache-p3 (-1)'' assay.

\begin{figure*}[t!]
\centering
    \includegraphics[trim = {0cm, 0cm, 0cm, 0cm}, clip, width=0.95\linewidth]{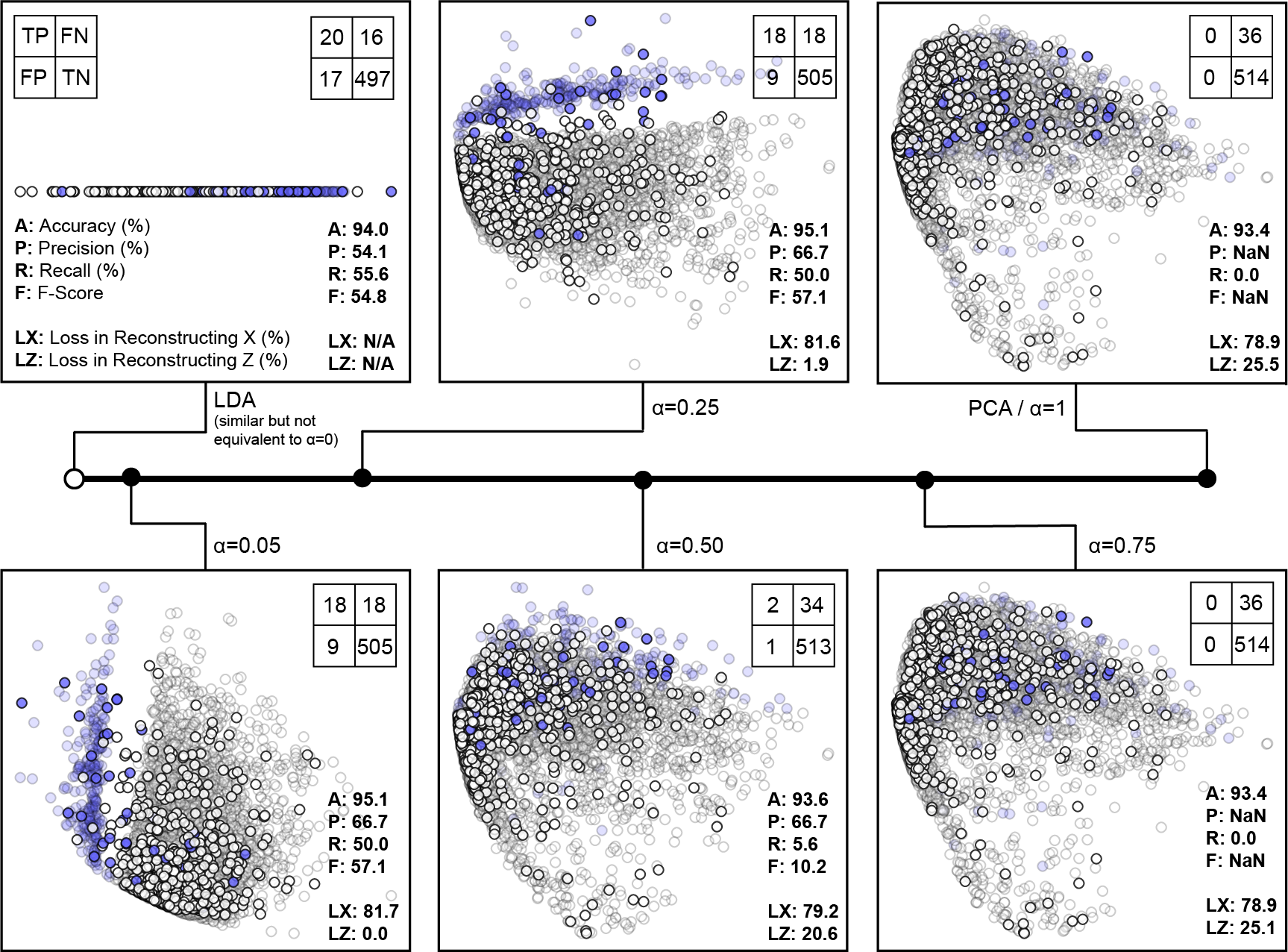}
\caption{\textbf{Effect of mixing parameter $\alpha$ on the 2D map and classification performance (on an out-of-sample dataset) for the ``tox21-ache-p3 (-1)'' assay.} White and blue points indicate ``non-toxic'' and ``toxic'' molecules, respectively. Opacity denotes train (translucent)/ test (opaque) split. In each panel is a confusion matrix showing the accuracy of logistic regression on the testing data as they appear in the resulting map, where ``TN'' indicates the number of ``true negatives'', ``FP'' indicates the number of ``false positives'', and so on. For comparison, logistic regression on the full-dimensional data obtains a confusion matrix of TP=17, TN=505, FP=9, FN=19.}
 \label{fig:pcovc_tox}\figspace
\end{figure*}

First, we compare the mapping results of PCovC as a function of our mixing parameter, $\alpha$, in Fig.~\ref{fig:pcovc_tox}. As expected, a PCA map of this dataset reflects little distinction between toxic and non-toxic molecules, where a subsequent analysis via logistic regression labels all data points as ``non-toxic,'' i.e., the majority population of the dataset. As shown in App.~D, nonlinear unsupervised dimensionality reduction algorithms such as t-SNE or UMAP also fail to distinguish toxic molecules from non-toxic molecules. As $\alpha$ is lowered and the classification task bears more weight in Eq.~\eqref{eq:pcovr_objective}, the activity cliff becomes clearer and our predictive power improves, even superceding that of logistic regression on the full-dimensional data.

For comparison, LDA produces a map very similar to PCovC at $\alpha=0$, wherein only the labels inform the projector matrix. This still performs better in subsequent classification tasks than PCA (and marginally worse than analogous PCovC maps). However, consider the implications of using a single-dimensional map to analyze the decisions made by a classification algorithm. This single-dimensional map can only reflect one dimension of chemical diversity, and all other informative dimensions are lost. Therefore, molecules that are similar in this first dimension, but otherwise dissimilar, cannot be distinguished. Furthermore, the orthogonal dimensions, such as the ``y'' axis in Fig.~\ref{fig:pcovc_tox} also carry scientific relevance -- these dimensions identify molecular characteristics that do \emph{not} influence the classification of the data points.

\begin{figure*}[t!]
\centering
    \includegraphics[trim = {0cm, 0cm, 0cm, 0cm}, clip, width=0.95\linewidth]{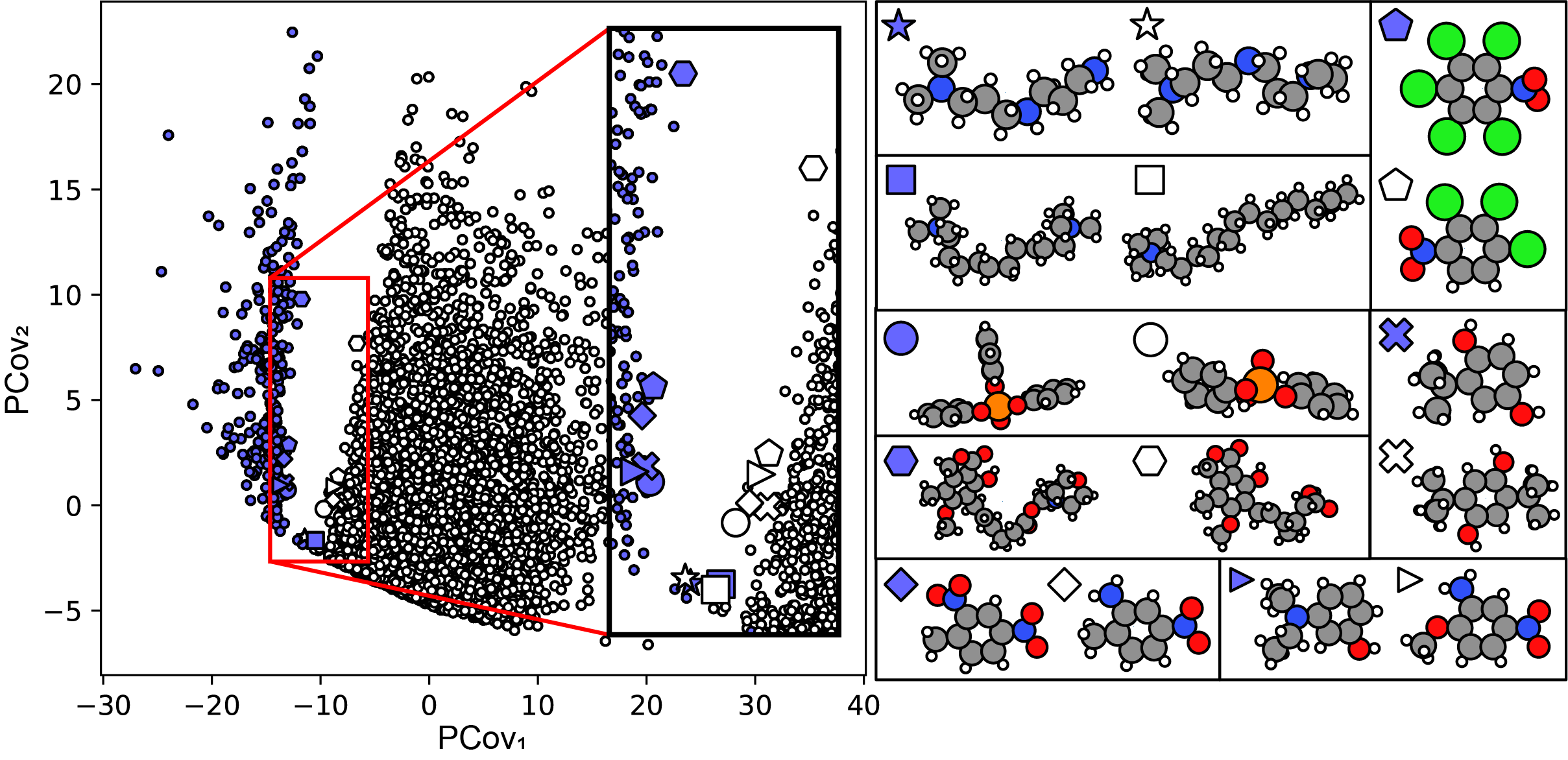}
\caption{\textbf{Molecules from the training set near the decision boundary} that demonstrate delineating features of toxicity based on the ``tox21-ache-p3 (-1)'' assay. The underlying map corresponds to PCovC at $\alpha=0.05$; markers denote different pairs of molecules, with blue and white markers corresponding to toxic and non-toxic molecules. Toxic/non-toxic molecule pairs were chosen to correspond to the molecules closest in PCovC space. Atoms are colored using the standardized Corey-Pauling-Koltun coloring scheme.}
 \label{fig:pcovc_tox_inset}\figspace
\end{figure*}

More concretely, consider the analysis of the map in Fig.~\ref{fig:pcovc_tox_inset}. To identify molecules along the decision boundary, we computed the distances between all ``toxic'' and ``non-toxic'' points in the first eight components of PCovC space and found the pairs closest to each other. The eight pairs shown demonstrate that \emph{the model has determined} that an activity cliff can appear due to small hydrocarbon functionalization, swapping of functional groups, or fluorination. While similar analysis can be achieved through visual inspection (e.g via \texttt{chemiscope.org} \cite{fraux_chemiscope_2020}), we identified these molecules through simple analysis of the PCovC latent space -- enabling higher throughput identification of boundary data points, which can then be used to identify new candidates for toxicity screening. 

\begin{figure*}[t!]
\centering
    \includegraphics[trim = {0cm, 0cm, 0cm, 0cm}, clip, width=\linewidth]{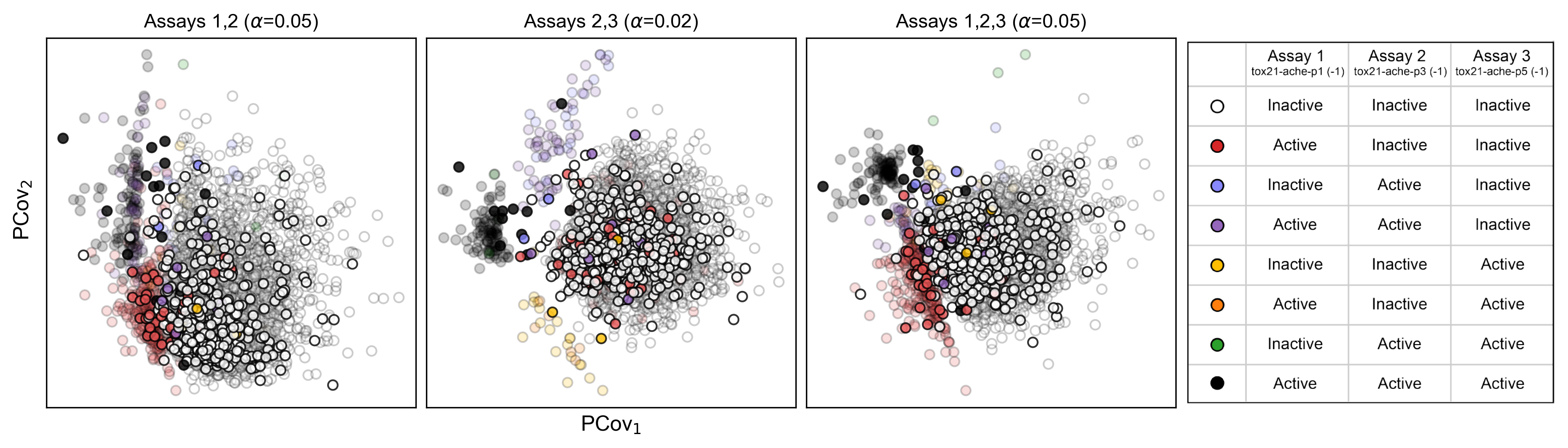}
\caption{\textbf{PCovC maps of the Tox21 dataset considering multiple assays} for classification, two double-assay maps (left, center) and a triple-assay map (right). Marker color corresponds to the values for the three assays, shown in the table to the right. Opacity denotes train (translucent)/ test (opaque) split.}
 \label{fig:pcovc_tox_multi}\figspace
\end{figure*}

Yet, drawing definitive conclusions about neurotoxicity from a single assay may be insufficient. By integrating data from multiple assays, we could strengthen conclusions on molecular design and determine confidence intervals. This constitutes a \textit{multitask} problem that other supervised dimensionality reduction algorithms like LDA are unable to handle. We denote the ``tox21-ache-p1 (-1)'' assay as ``Assay 1'', the ``tox21-ache-p3 (-1)'' assay (previously discussed) as ``Assay 2'' and the ``tox21-ache-p5 (-1)'' assay as ``Assay 3''. We can generate multitask PCovC maps, seen in Fig. \ref{fig:pcovc_tox_multi}, to incorporate information from each of these assays into a single map. 

Of the three assays, Assay 1 is the most sensitive indicator of toxicity. In our postprocessed dataset, 488 of the 5,767 (8.46\%) compounds are labeled as toxic by this assay, compared to 193 (3.35\%) and 146 (2.53\%) compounds for the Assays 2 and 3, respectively. This means that there are significantly many more data points considered ``toxic'' by Assay 1 that are ``non-toxic'' according to Assays 2 and 3. The PCovC maps for these singular assays and the other assay combinations not shown in Fig.~\ref{fig:pcovc_tox_multi} are shown in the Supplementary Information. Note the number of red points, indicating molecules only considered ``toxic'' by Assay 1, and black points, indicating molecules considered toxic by all assays. A map generated from solely Assay 2 positions these red points with the ``non-toxic'' white points (as shown in the Supplementary Information). The clearest separation occurs in the map built on Assays 1 and 2, which delineates both decision boundaries clearly and discretely. One can imagine doing similar analysis to that of Fig.~\ref{fig:pcovc_tox_inset} to identify the molecules along these boundaries to determine the differentiation by the two assays. 

\subsection{Parsing multi-class datasets: spectral analysis of organosulfur bonding environments}

One key limitation of using purely unsupervised models for QSPR is their tendency, by design, to map only the highest-variance or highest-information dimensions. However, many datasets contain multiple splits of interest corresponding to different properties.
We apply PCovC to X-ray absorption near edge structure (XANES) spectra that are used to characterize the types of bonds that may occur in organosulfur compounts. We use the dataset published in \cite{tetef_unsupervised_2021}, which contains 769 XANES spectra of organosulfur molecules, and consider three ``levels'' of property classification: sulfur oxidation state (-2, 0, +2), type of sulfur bond (sulfide, thiocarbonyl, thiol/mercaptan, sulfoxide, and sulfone), and whether the sulfur is the member of a conjugated system (aromatic) or not (aliphatic). Note that the classifications by oxidation state and bond type are \textit{multiclass}, thus, in these cases LDA can be used to generate a multidimensional latent space.

\begin{figure*}[t!]
\centering
    \includegraphics[trim = {0cm, 0cm, 0cm, 0cm}, clip, width=\linewidth]{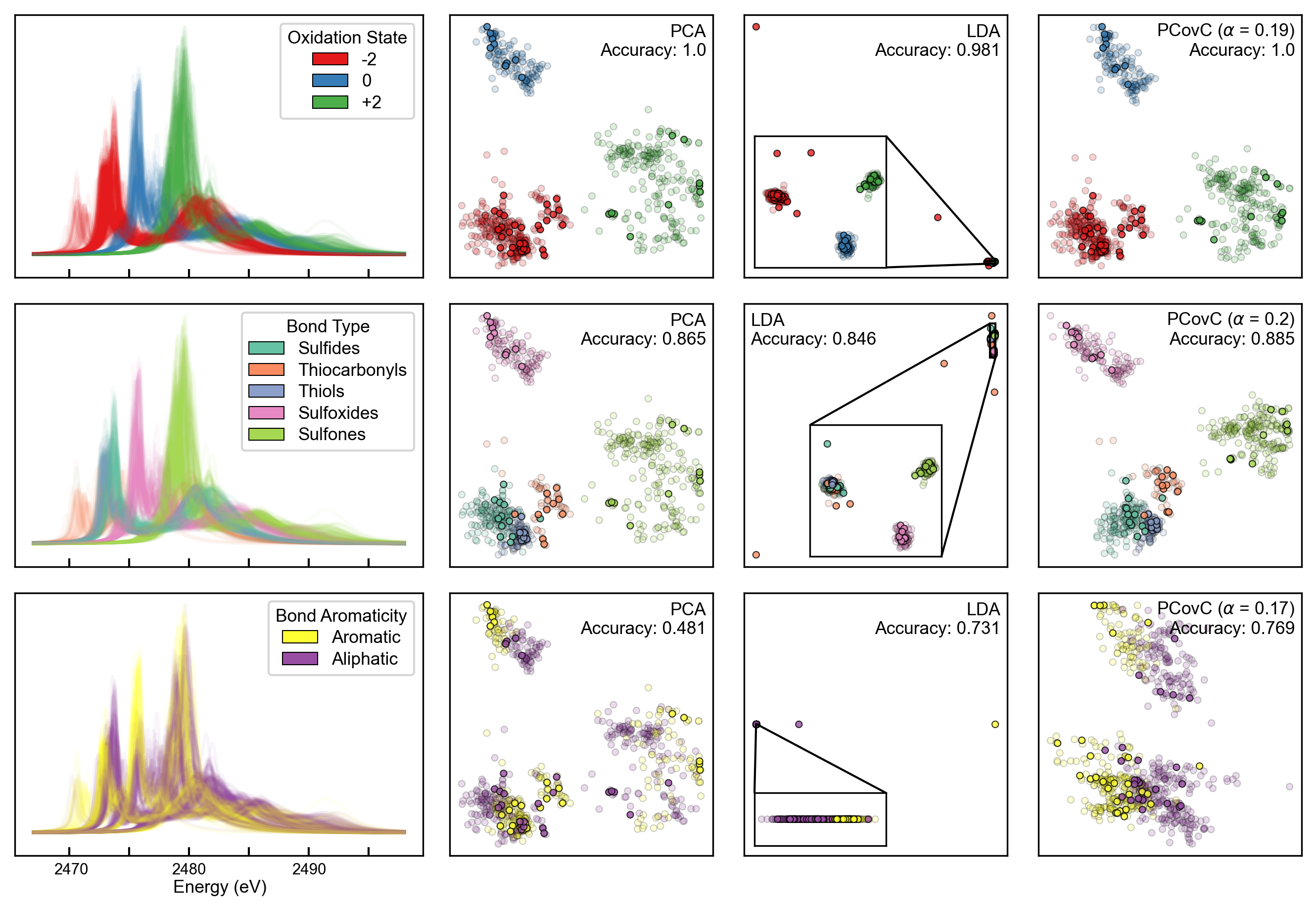}
\caption{\textbf{Analysis of organosulfur dataset for different classification targets.} Rows correspond to classification based on oxidation state (top), bond type (center), and aromaticity (bottom). (left) Underlying XANES spectra, colored by the class delineations. The main figure corresponds to the underlying spectra. (right) Maps according to the first and second learned PCA, LDA, and PCovC dimensions, colored by the class delineations. The PCA map will be the same for all three analyses. In the upper right, we indicate the accuracy of a logistic regression model on the resulting 2D map for the offset testing data.}
 \label{fig:pcovc_sulfur}\figspace
\end{figure*}

As can be seen in the first row of Fig.~\ref{fig:pcovc_sulfur}, it is straightforward to identify the the signals corresponding to different oxidation states, as the data separates well based on this set of classes. Because of this, a logistic regression model trained on the 2D PCA latent space yields perfect test accuracy, and LDA only misclassifies one test compound. Because PCA is well-suited to separate oxidation states, the choice of $\alpha$ for a PCovC map is somewhat arbitrary, with $\alpha=0.19$ minimizing the PCovC loss. Note that the map shown is qualitatively similar to the PCA projection, with minute differences.

Bond types are not well-separated in the PCA projection, given that the -2 oxidation state contains three different types. As shown in Fig. \ref{fig:pcovc_sulfur}, the sulfides and the thiols are the most difficult to separate, leading to a loss in accuracy for both PCA (87\%) and LDA (85\%) . However, a logistic regression model trained on the PCovC latent space is still able to achieve a 89\% bond type test classification accuracy, despite the fact that the $\alpha$ value was not chosen to maximize it. This demonstrates how incorporating classification loss into our mapping procedures ``refocuses'' our model; in purely unsupervised schema, having a wealth of data within one class versus another leads to this data dominating the mapping task. 

The most difficult task is separating the compounds by aromaticity, as this is a property that cross-cuts the ``higher-order'' classes. As is suggested by the maps in Fig.~\ref{fig:pcovc_sulfur}, the principal components seem to be capturing the oxidation state and bond type of the organosulfur compounds. However, because neither of these properties has a direct relationship with bond aromaticity, a logistic regression model trained from the PCA latent space struggles (accuracy of 48\%). However, PCovC and LDA are each able to find latent spaces in which the compounds are well delineated by their aromaticity, with accuracies of 77\% and 73\%, respectively. The PCovC latent space also captures the difference in bond type via the second principal covariate, where the upper cluster of points correspond to the sulfones, and the lower cluster encompasses the sulfoxides, thiocarbonyls, sulfides, and thiols.

\begin{figure*}[t!]
\centering
    \includegraphics[trim = {1.5cm, 0cm, 1.5cm, 0.6cm}, clip, width=\linewidth]{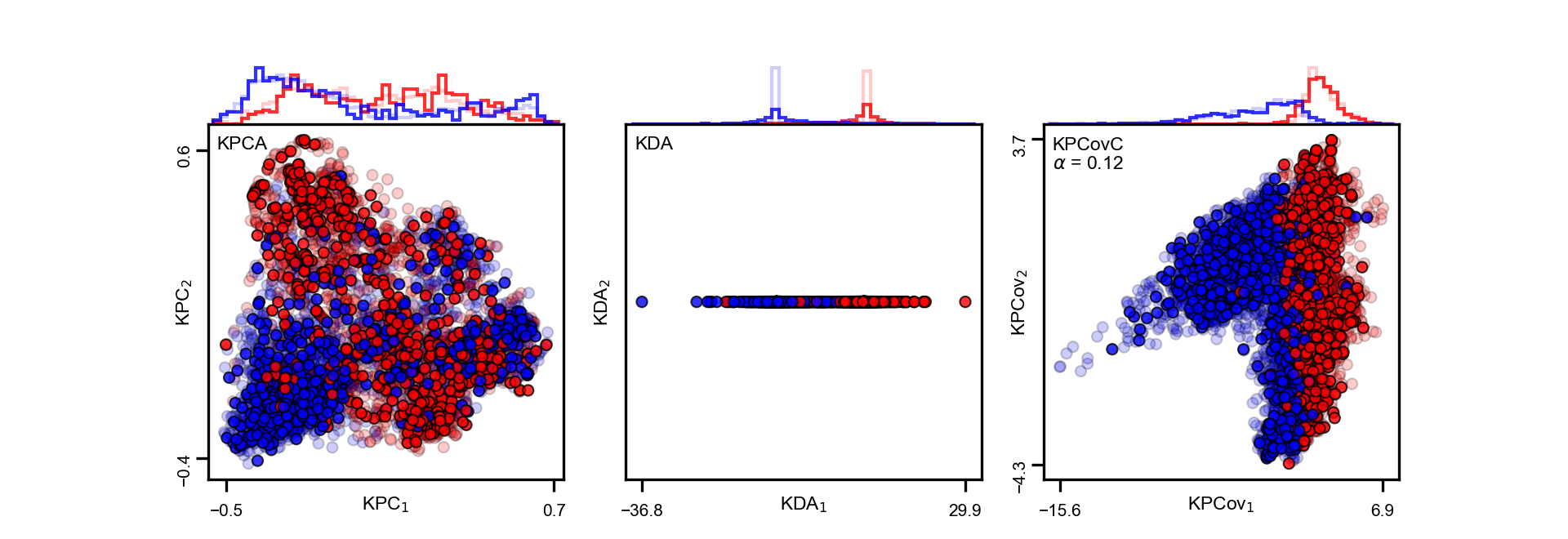}
\caption{\textbf{Projections of inorganic materials found in \cite{zhuo_predicting_2018} using Kernel PCA, KDA, and Kernel PCovC.} Color and opacity denote metal (blue) versus non-metal (red) and training (transparent) versus testing (opaque) splits, respectively. Above each map is the distribution of points in each class.}

 \label{fig:pcovc_bandgap_distributions}\figspace
\end{figure*}

So, why was PCovC able to consistently outperform LDA, an algorithm \textit{designed} to find latent spaces that best delineate classes? The first step in computing any LDA projection involves finding and projecting data to the lowest dimensional subspace shared by the set of class means. In doing so, any directions of the data orthogonal to this subspace are lost, which can result in LDA dropping important intra-class scatter. This loss of information manifests itself via a hyperconcentration of points within their classes, which results in LDA overfitting the training data and struggling to correctly predict the bonding environments of out-of-sample chemistries. With PCovC, on the other hand, there is explicit regularization within the classification procedures, and we can decide how much inter- and intra-class variance we wish to retain in our map through tuning $\alpha$.

\subsection{Understanding feature importance in nonlinear classification of metals/nonmetals}

Finally, what if each of our features are physically meaningful but we require nonlinear models for effective classification? In this instance, it is unclear how these meaningful features contribute to the learned weights of a classification algorithm. However, with robust structure-property maps, we can still interpret these nonlinear feature relationships. In this case study, we apply a Kernelized PCovC (KPCovC) on the dataset of 4,916 inorganic materials published in \cite{zhuo_predicting_2018} to understand the relationships between elemental properties and metallic behavior. Each of these materials is represented as a set of 136 properties of the constituent elements, and the learnable target is whether the material is a metal or a non-metal. Following the procedure in \cite{zhuo_predicting_2018}, we define a non-metal and metal as materials that do or do not have an electronic band gap, and extract the evidence matrix from a trained support vector classification (SVC) model, where $\mathbf{Z}\equiv \mathbf{KP}_{KZ},$ where $\mathbf{K}$ is a similarity matrix determined by a radial basis function (RBF) kernel, using a Gaussian width $\gamma$ of 0.01.

With this kernel matrix, we compute the Kernel PCA, Kernel Discriminant Analysis (KDA) and KPCovC projections of the elemental property data, shown in Fig.~\ref{fig:pcovc_bandgap_distributions}. As was shown in \cite{zhuo_predicting_2018}, we also observe that applying KPCovC with a support vector machine classifier optimizes classification performance and best exposes the decision boundary; using KPCovC with $\alpha=0.12$, the accuracy to 90\% compared to 88\% and 65\% with the analogous KDA and KPCA, respectively.

\begin{figure*}[t!]
\centering
    \includegraphics[trim = {0cm, 0cm, 0cm, 0cm}, clip, width=\linewidth]{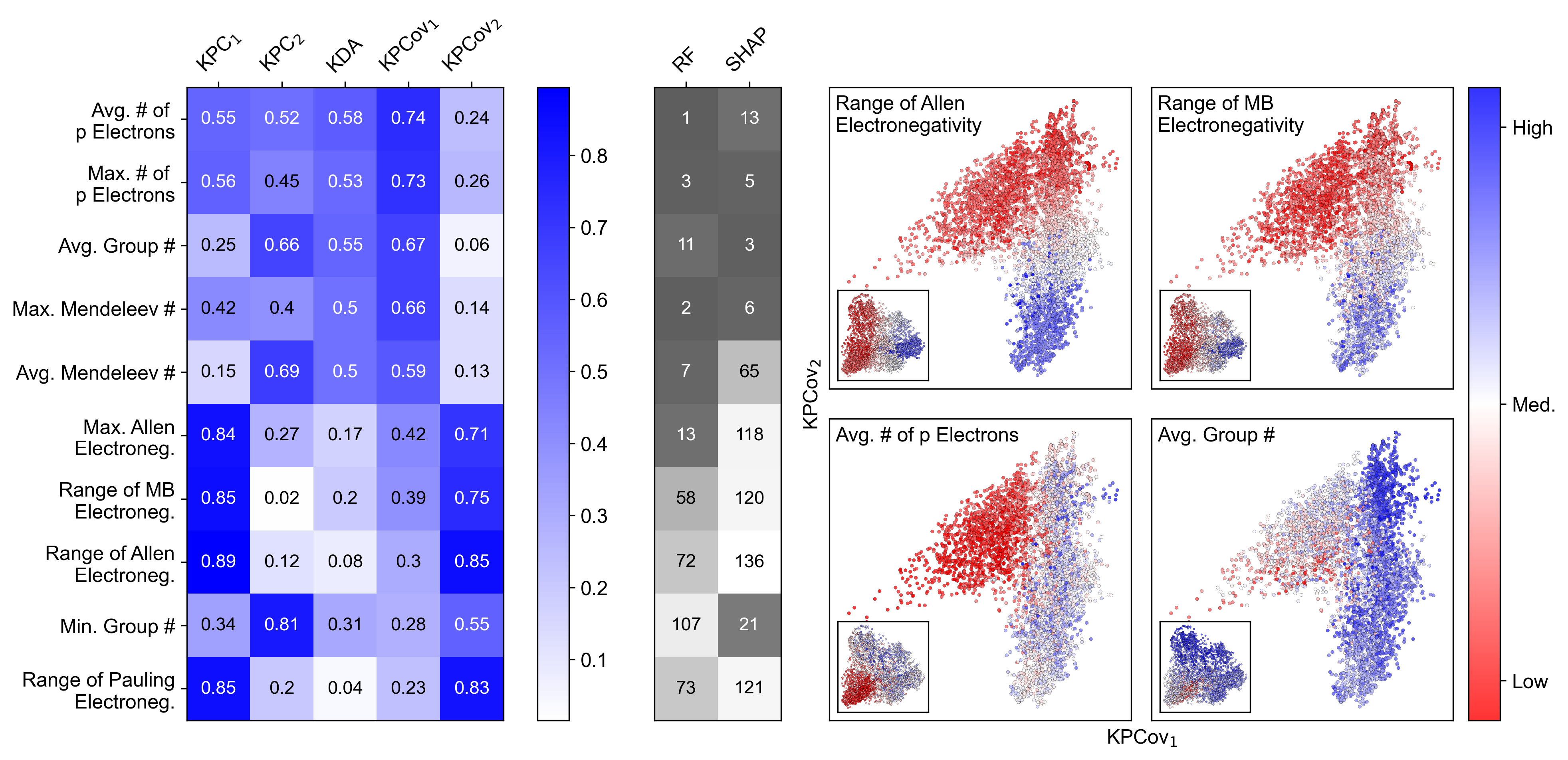}
\caption{\textbf{Correlations of materials characteristics with the nonlinear KPCovC and KPCA maps.} (left) Absolute value of Pearson correlation coefficients for different input features with the resulting maps from Fig. \ref{fig:pcovc_bandgap_distributions}. Darker squares denote stronger correlation. (middle) Ranked SHAP values and random forest (RF) feature importance. Darker squares indicate a higher importance.} (right) Recoloring of KPCovC (main) and KPCA (inset) by feature values, from low (red) to high (blue) on a linear scale.
 \label{fig:pcovc_bandgap_correlations}\figspace
\end{figure*}

With the KPCovC map in Fig.~\ref{fig:pcovc_bandgap_distributions}, we could again perform analysis by computing distances and identifying boundary data points in the KPCovC latent space similar to Fig.~\ref{fig:pcovc_tox_inset}. However, given that the features have physical significance, we instead propose to view the map from the lens of the feature set itself. More specifically, we can quantitatively determine the discriminating features in the data by computing the Pearson correlation coefficient between each feature and the first and second principal covariates. In Fig.~\ref{fig:pcovc_bandgap_correlations}, we demonstrate such an analysis, while also computing correlations between other latent space dimensions. We find that features based on the electronegativity and ionization energy of the constituent elements, while highly correlated with the first principal component, have only a moderate to weak correlation with the first principal covariate. This indicates that, while electronegativity and ionization energy objectively demonstrate the greatest variance in the data, our model does not associate either property with changes in metallic behavior (upper right panels of Fig.~\ref{fig:pcovc_bandgap_correlations}). The same conclusion can be reached by viewing the correlation with the KDA 1D projection, but in the case of KDA, the information these features provide is mostly discarded on the map. On the other hand, the KPCovC map retains this orthogonal information (as shown by their strong correlations with the second principal covariate) while also highlighting that they largely do not contribute to the class outcome. Instead, as shown in the lower right panels of Fig.~\ref{fig:pcovc_bandgap_correlations}, features related to the number of p electrons and Mendeleev number, which are not strongly correlated with the first principal component, show strong correlation with the first principal covariate.

In general, the SHAP values and random forest feature importances correlate strongly with the representation of the features in the first principal covariate. However, interpreting models using these techniques has limitations. First, for these feature importance metrics to be useful, it is necessary that the features themselves can be interpreted. While this is not the case here, when abstract descriptors are used, such as the RDKit fingerprint in Sec.~\ref{sec:tox}, it will be difficult to relate these metrics to chemical concepts, which will make QSPR unachievable. With PCovC, model interpretation does not have to be limited to features the model uses for training, as visualizing or computing correlation coefficients can be performed trivially with the PCovC latent space. In addition, it is difficult to determine the span of the feature's importance across the space of observations using solely feature importance metrics. For example, consider the map showing the average number of p electrons in the KPCovC projection in \ref{fig:pcovc_bandgap_correlations}. This map shows that there is a strong population of metals with a small number of p electrons, but this population does not span the entire population of metals. While SHAP can similarly achieve sample-level resolution in this manner, PCovC can also be used to identify subgroups \textit{across} the decision boundary, such as the sets of materials with high and low electronegativities in Fig. \ref{fig:pcovc_bandgap_correlations}.

\subsection{Enhancing nonlinear dimensionality reduction techniques with MNIST}

This slew of methods, as demonstrated in \cite{yang_dimensionality_2021, cristian_diffusion_2024}, extend in usefulness beyond those problems that are chemically-motivated. Furthermore, they offer benefits to traditional mapping workflows. To demonstrate both of these points, we conclude with a two-step workflow on the benchmark dataset MNIST.

\begin{figure*}[t!]
\centering
    \includegraphics[trim = {0cm, 11.8cm, 0cm, 0cm}, clip, width=\linewidth]{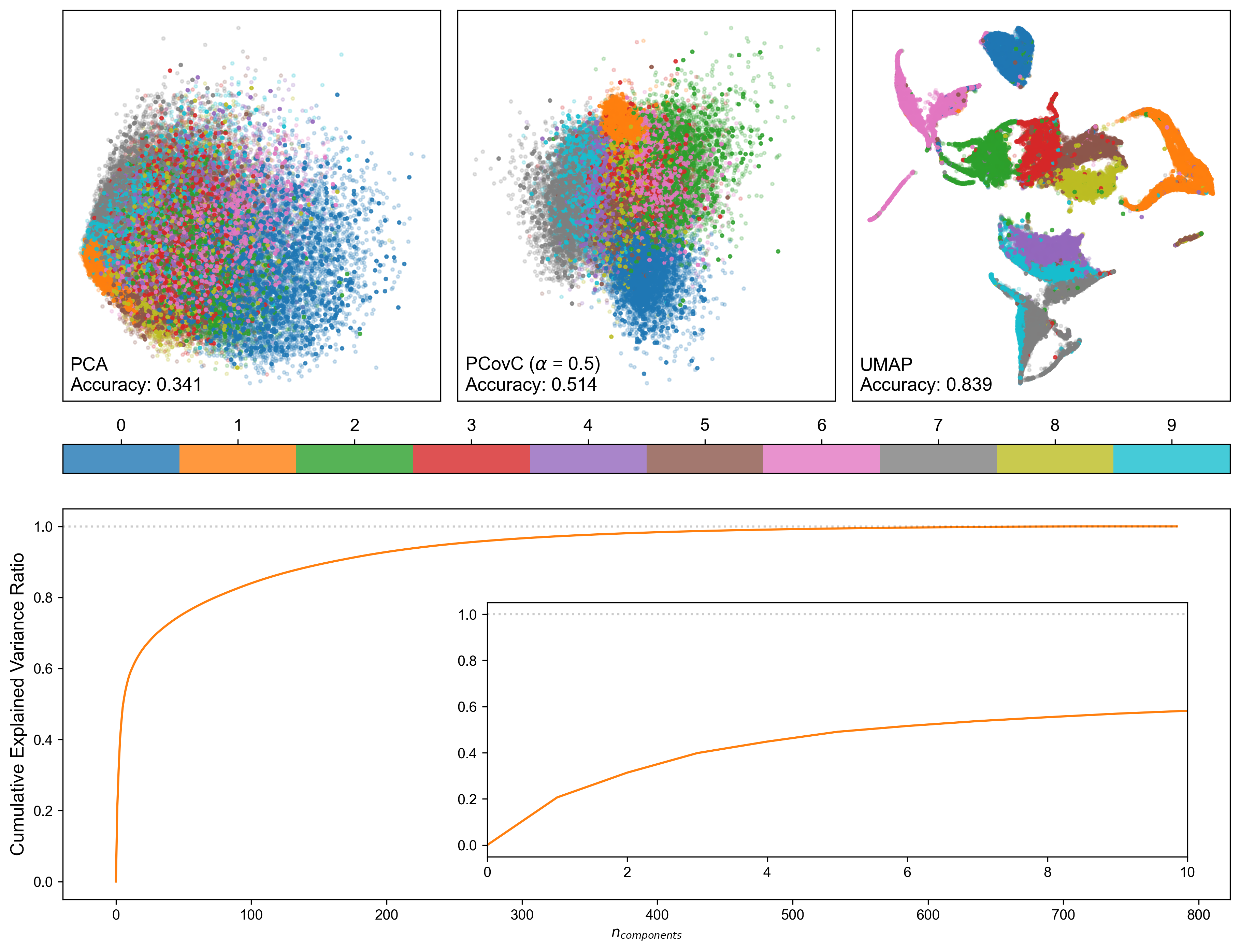}
\caption{\textbf{Comparison between PCA, PCovC, and UMAP on the MNIST dataset.} (top row) Maps according to the first and second PCA, PCovC, and UMAP dimensions. In the lower left, we denote the accuracy of a logistic regression model on the resulting 2D map for the offset testing data. Color corresponds to the digit label, and opacity corresponds to the training (translucent) versus testing (opaque) split.}
 \label{fig:pcovc_vs_pca_MNIST}\figspace
\end{figure*}

Nonlinear methods are often essential when facing problems in which local correlations between features dictate class labels. Consequently, with the MNIST handwritten digits dataset, dimensionality reduction algorithms that optimize the preservation of local neighborhoods of points in their reduced spaces, such as t-SNE and UMAP, can find embeddings with far cleaner digit separation than linear models \cite{mcinnes_umap_2020}. With MNIST, a logistic regression model trained on the flattened images can achieve an accuracy of 92\%. However, as can be seen in Fig. \ref{fig:pcovc_vs_pca_MNIST}, UMAP significantly outperforms PCovC in a latent space linear classification task, despite the fact that UMAP is both unsupervised and does not enforce linear cluster separation. This failure occurs because, while linear classification is possible in the high-dimensional space, the locally defined class structure requires more than two dimensions to be resolved by PCovC's linear projection. This advantage, however, is balanced by an increased computational cost. For this reason, when working with large datasets, many have considered using PCA as a preprocessing step to reduce data and make nonlinear dimensionality reduction feasible \cite{yang_dimensionality_2021, cristian_diffusion_2024}. This begs the question: if there exists linearly decodable information that can be used to \textit{partially} separate classes, can this information be used to enhance a nonlinear mapping strategy?

\begin{figure*}[t!]
\centering
    \includegraphics[trim = {0cm, 0cm, 0cm, 0cm}, clip, width=\linewidth]{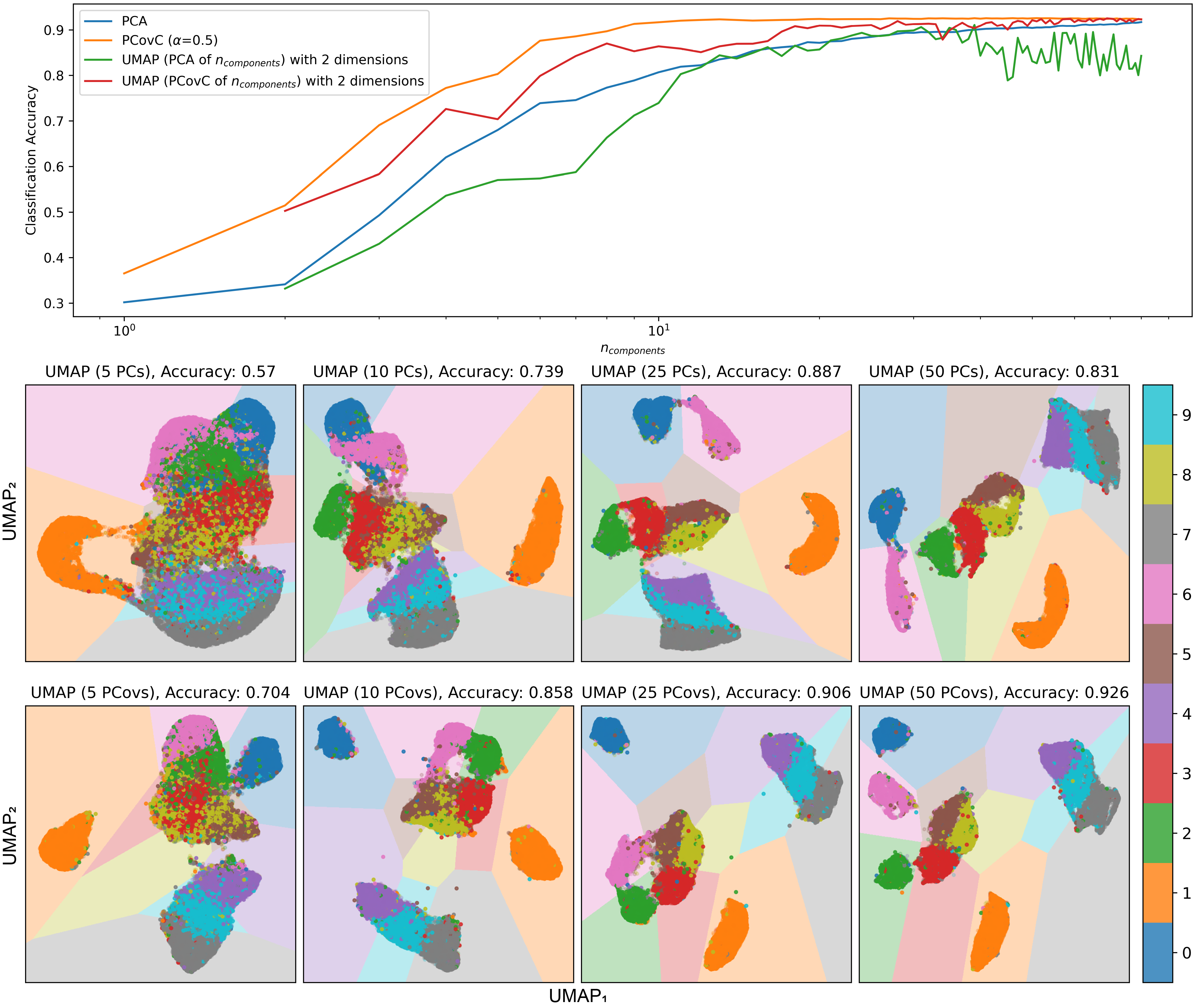}
\caption{\textbf{UMAP projections on the MNIST dataset with PCovC and PCA pre-processing.} (top) Accuracy of a logistic regression model trained in PCA, PCovC, 2D UMAP on PCovC components, and 2D UMAP on PCA components. (middle row) UMAP projections of MNIST obtained from PCA-reduced latent spaces. (bottom row) UMAP projections of MNIST obtained from PCovC-reduced latent spaces. In the titles, we indicate the accuracy of a logistic regression model on the resulting 2D map for the testing data. Color corresponds to the digit label and the backgrounds show the estimated decision boundaries.}
 \label{fig:pcovc_MNIST}\figspace
\end{figure*}

In Fig. \ref{fig:pcovc_MNIST}, we explore the use of PCovC as a replacement to PCA as a preprocessing step to UMAP. We observe that UMAP projections learned from PCovC-reduced subspaces of the data result in clearer digit separation than those learned from both the PCA-reduced subspaces and UMAP on the high-dimensional space. In practice, this means that PCovC can be used in two ways: PCovC can substitute PCA at a fixed subspace dimensionality to enhance the quality of the subsequent UMAP embedding without a significant increase in the computational cost, or it can be used to achieve stronger dimensionality reduction prior to UMAP without sacrificing class delineation in the resulting projection.

\section{Conclusions}

In this work, we demonstrated that hybrid supervised-unsupervised mapping techniques, namely principal covariates classification, can be transformative tools for visualizing data spaces through the lens of classification problems. The corresponding open-source code is available at \url{https://github.com/cersonsky-lab/PCovC}.
We observed that, compared to PCA and LDA, combining classification performance and information preservation in a single mapping strategy can unveil decision boundaries while still preserving similarity between data points, and can even lead to improvements in classification performance. In computational chemical analysis, these aspects can lead to furthering our ability to identify and interpret structure-property relationships. 

\bibliography{references}
\clearpage

\appendix 

\section{Detailed methods used to obtain results in the main text}

\subsection{Principal Covariates Classification}

PCovC is implemented in the open-source package \texttt{scikit-matter} \cite{goscinski_scikit-matter_2023}, a \texttt{scikit-learn}-affiliated package focused on methods born out of chemical sciences. This package follows the same API as \texttt{scikit-learn} \cite{pedregosa_scikit-learn_2011}, and thus obtaining a PCovC map follows the same \texttt{fit} and \texttt{transform} procedure similar to obtaining PCA in \texttt{scikit-learn}. 

Behind the curtain, obtaining a PCovC map requires a few key steps. First, we must obtain $\mathbf{Z}$ for use in Eq.~\eqref{eq:pcovc_gram} through fitting a linear classifier with $\mathbf{X}$ and $\mathbf{Y}$. The exact choice of linear classification model is flexible. The projector from the feature space to the latent space can be mathematically derived through following the procedure in \cite{helfrecht_structure-property_2020}, but with the evidence matrix $\mathbf{Z}$ used instead of the linear regression approximation $\mathbf{\hat{Y}}$. When $n_{\mathbf{features}}>n_{\mathbf{samples}}$, one should obtain the projectors by performing an eigendecomposition of a modified Gram matrix $\mathbf{\tilde{K}}$ (Eq.~ \ref{eq:sample_pcovc}).

\begin{subequations}
\label{eq:sample_pcovc}
\begin{gather}
\mathbf{P}_{XT} = \left( \alpha \mathbf{X}^T + (1 - \alpha) \mathbf{W} \mathbf{Z}^T \right) \mathbf{U}_{\tilde{\mathbf{K}}} \mathbf{\Lambda}_{\tilde{\mathbf{K}}}^{-\frac{1}{2}} \label{eq:sample_pcovr_pxt} \\
\mathbf{P}_{TX} = \mathbf{\Lambda}_{\tilde{\mathbf{K}}}^{-\frac{1}{2}} \mathbf{U}_{\tilde{\mathbf{K}}}^T \mathbf{X} \label{eq:sample_pcovr_ptx}
\end{gather}
\end{subequations}
In Eq.~\eqref{eq:sample_pcovc}, $\mathbf{U}_{\mathbf{\tilde{K}}}$ and $\mathbf{\Lambda}_{\mathbf{\tilde{K}}}$ are obtained through performing an eigendecomposition of $\mathbf{\tilde{K}}$, and $\mathbf{W}$ represents the set of weights from the linear classifier fit between $\mathbf{X}$ and $\mathbf{Y}$. When $n_{\text{features}}<n_{\text{samples}}$, it is instead faster to use a modified covariance matrix $\mathbf{\tilde{C}}$ (Eq.~\eqref{eq:feature_pcovc}).

\begin{subequations}
\label{eq:feature_pcovc}
\begin{gather}
\mathbf{P}_{XT} = \left( \mathbf{X}^T \mathbf{X} \right)^{-\frac{1}{2}} \mathbf{U}_{\mathbf{\tilde{C}}}^T \mathbf{\Lambda}_{\mathbf{\tilde{C}}}^{\frac{1}{2}}\label{eq:feature_pcovc_pxt}\\
\mathbf{P}_{TX} = \mathbf{\Lambda}_{\mathbf{\tilde{C}}}^{-\frac{1}{2}} \mathbf{U}_{\mathbf{\tilde{C}}}^T \left( \mathbf{X}^T \mathbf{X} \right)^{\frac{1}{2}} \label{eq:feature_pcovc_ptx} 
\end{gather}
\end{subequations}

Likewise, $\mathbf{U}_{\mathbf{\tilde{C}}}$ and  $\mathbf{\Lambda}_{\mathbf{\tilde{C}}}$ are obtained from performing an eigendecomposition of $\mathbf{\tilde{C}}$.
    
Finally, we require $\mathbf{P}_{TZ}$, a projector that can map the latent space and the class likelihoods. This projector can be found by extracting the weights from a linear fit between $\mathbf{T}$ and $\mathbf{Z}$. To obtain predicted class labels, one can similarly apply an activation function to $\mathbf{TP}_{TZ}$.

Selecting $\alpha=1$ will still result in a standard PCA decomposition; selecting $\alpha=0$, on the other hand, can result in re-learning $\mathbf{P}_{TZ}$ when the dimensionality of $\mathbf{T}$ is the same as the dimensionality of $\mathbf{Z}$.

\subsubsection{Computational Scaling of PCovC}
The primary cost of performing PCovC lies in the calculation and subsequent eigendecomposition of the modified Gram matrix $\mathbf{\tilde{K}}$ or modified Covariance matrix $\mathbf{\tilde{C}}$. For a feature matrix $\mathbf{X}$ of dimension $m \times n$, the explicit calculation of $\mathbf{\tilde{K}}$ or $\mathbf{\tilde{C}}$ will have complexity $\bigO(m^2n)$ or $\bigO(mn^2)$, respectively. The equivalence of their eigendecompositions implies this can be done on the option corresponding to the lesser of $m$ and $n$. The scaling of the eigendecomposition is $\bigO(kmn)$, where $k$ is the number of eigenvectors we choose to retain. The overall complexity for PCovC will therefore be $\bigO(mn^2 + kmn)$, where we note that the hidden prefactor for the eigendecomposition is typically much larger than that of matrix multiplication due to the latter's parallelizability.

\subsubsection{Inner Product of Evidence Tensors}

Depending on the problem of interest, it may be necessary to perform PCovC with a dataset that involves multiclass, multilabel classification. In these instances, it will be necessary to compute $\mathbf{ZZ}^T$ with a 3D $\mathbf{Z}$. This can be done following Eq.~\eqref{eq:pcovc_Z}.

\begin{equation}
\label{eq:pcovc_Z}
    (\mathbf{ZZ}^T)_{ij} = \sum_{c=1}^{n_{\text{classes}}}\sum_{l=1}^{n_\text{labels}}\mathbf{Z}[i, c, l]\cdot \mathbf{Z}[j, c, l]
\end{equation}

This approach can also be extended to problems in which each label corresponds to a different amount of classes, as $n_{\text{classes}}$ can be treated as a variable that changes across labels.

\subsection{Datasets}
\label{app:datasets}

The three datasets used in this work have been published in previous studies.

\subsubsection{Tox21}

This dataset, released in \cite{wu_trade-off_2021}, contains Tox21 \cite{richard_tox21_2021} bioassay data for 68 end points and 7,660 compounds, which are provided as SMILES strings. To better enable a visual analysis of the compounds, we separately convert the compounds into ASE \cite{hjorth_larsen_atomic_2017} objects using RDKit \cite{landrum_rdkitrdkit_2025}; 10 compounds were dropped from the dataset as they resulted in a ``bad conformer id'' error from RDKit. We then follow the preprocessing scheme in \cite{wu_trade-off_2021}, in which all compounds other than those with an ``inactive'' result or the most common active results are not considered for model fitting. For the three acetylcholinesterase assays considered in this study, the number of compounds labeled as active antagonists outnumbered those that were labeled as active agonists by a wide margin. Thus, the active antagonists were kept and the active agonists were discarded. We use the same train-test split provided in the dataset itself; for the rigorous analysis of the ``tox21-ache-p3 (-1)'' assay, the result is a dataset of 6,357 compounds (5,807 train, 550 test). For the multitask PCovC maps, because we discard chemicals that are not inactive or active antagonists for all three assays, the dataset consists of 5,767 compounds (5,277 train, 490 test). As in the original paper, we then convert the SMILES strings of each of the compounds into RDKit fingerprints with default fingerprint size (2,048 bits).

\subsubsection{Sulfur Spectra}

In this case study, we analyze the dataset, released in \cite{tetef_unsupervised_2021}, that contains 769 sulforganic compounds and their corresponding XANES spectra, each having been computed using time-dependent density functional theory. We use the same train-test split as was used in the original paper for each of the three classification tasks, which results in a training dataset of 717 spectra and a test dataset of 52 spectra (6.8\%). We follow the same steps for preprocessing as was done in the original paper, in which we first divide each spectra by the highest intensity value in the training dataset, then mean-center each feature.

\subsubsection{Metal/Nonmetal Classification}

The dataset used for metal/nonmetal classification was published in \cite{zhuo_predicting_2018}. It contains a set of 4,916 inorganic materials, each described by the average, difference, maximum, and minimum of 34 properties of its respective constituent elements. Following the procedure in \cite{zhuo_predicting_2018}, we perform an 80\%-20\% train-test split (3932 train, 984 test) and scale each feature in the data to have zero mean and unit variance. 

\subsubsection{MNIST}

The MNIST dataset \cite{lecun_mnist_2010} was obtained via the tensorflow API. MNIST contains a training set of 70,000 handwritten digits (60,000 train, 10,000 test) in the form of 28x28 grayscale images. Before applying dimensionality reduction, we first flatten each of the images to form a feature vector of length 784, then scale the intensity of each pixel in the image to have zero mean and unit variance.

\section{Theory underlying existing methods for data visualization and explainability}
\label{app:methods}

\subsection{Linear and Kernel Principal Components Analysis}

PCA is a widely-used linear dimensionality reduction technique that aims to retain variance, or information, present in the feature matrix $\mathbf{X}$. More specifically, it aims to find the orthonormal projection $\mathbf{T}=\mathbf{XP}_{XT}$ that minimizes the reconstruction error $l$ of using the low-dimensional projection to reconstruct the original feature matrix:

\begin{equation}
\ell_{proj} = ||\mathbf{X-TP}_{TX}||^2/n_\text{{samples}}
\end{equation}

It is worth noting that $\mathbf{P}_{TX}$ can be found by invoking the requirement that $\mathbf{P}_{XT}$ is orthonormal: $\mathbf{P}_{TX} = \mathbf{P}_{XT}^T$. Additionally, a closed-form solution for $\mathbf{P}_{XT}$ exists, which can obtained through performing an eigendecomposition of the covariance matrix. PCA can be very useful for understanding and visualizing what a dataset may look like, however, it does not explicitly retain discriminative information, so the resulting latent space may not be useful if the directions in the data that are responsible for delineating classes are not the same as those that contain the most variance.

\subsection{Linear Discriminant Analysis}
\label{app:lda}
Linear discriminant analysis (LDA) is a well-known classifier that can also perform \textit{supervised} dimensionality reduction \cite{hastie_elements_2009}. Mathematically, LDA involves modeling the density of each class $k$ as a multivariate Gaussian:

\begin{equation}
    f_k(x)=\frac{1}{(2\pi)^{n_\text{{samples}}/2}|\Sigma_k|^{1/2}}e^{-\frac{1}{2}(x-\mu_k)^T\Sigma_k^{-1}(x-\mu_k)}
\end{equation}

For linear discriminant analysis, we assume that classes have the same covariance matrix $\Sigma_k=\Sigma$, which leads to the following linear discriminant function:

\begin{equation}
    \log P(y=k|x) = x^T \Sigma^{-1} \mu_k - \frac{1}{2} \mu_k^T \Sigma^{-1} \mu_k + \log P(y=k)
\end{equation}

For dimensionality reduction, LDA will identify a linear projection that maximizes the separation between classes. One can consider the $n_{\text{classes}}$ class means as vectors in $\mathcal{R}^{n_{\text{features}}}$. There exists some subspace $\mathbf{H}$ that contains each of these vectors that is of dimension at most $n_{\text{classes}}-1$. Directions in the data orthogonal to $\mathbf{H}$ contribute to each class equally and thus do not contribute in distinguishing classes. Thus, computing Euclidean distances in $\mathbf{H}$ will allow us to reduce the data to dimension $n_{\text{classes}}-1$ without sacrificing any information useful for classification. To obtain a lower-dimensional latent space optimal for classification, one can project to a linear subspace $\mathbf{H_L}$ that maximizes the variance of the class means after projection. $\mathbf{H_L}$ can be obtained through performing PCA on the class means. LDA dimensionality reduction can be useful because it will preserve the inter-class variance in the data, which is useful in discriminating between classes. However, in doing so, the intra-class variance is lost, restricting the ability to understanding relationships between points in the same class.

\subsection{SHAP Values and Saliency Maps}

SHAP is a framework for interpreting machine learning model predictions that draws from Shapley values in cooperative game theory. SHAP assigns each feature a score (or SHAP score) that quantifies the change in the conditional expectation of a model that would occur by including the feature, while also accounting for all possible permutations of features. The SHAP value for a feature $i$ is given by:

\begin{widetext}
\begin{equation}
\mathbf{SHAP}_i = \sum_{S \subseteq F \setminus i} \frac{|S|!(n_{\text{features}} - |S| - 1)!}{n_{\text{features}}!} \ E[f(z)|z_S] - E[f(z \setminus i)|z_{S \setminus i}]
\label{eq:shapley_value}
\end{equation}
\end{widetext}

In Eq.~\eqref{eq:shapley_value}, $S$ represents a subset of the feature set $F$ that does not include $i$, and $z_S$ represents a vector of features including $S$. SHAP is useful because it can assign quantitative, robust feature importance values for each feature regardless of model architecture, which can then be used to either interpret model predictions or perform feature selection. However, SHAP scales poorly when $n_{\text{features}}$ becomes large \cite{arenas_tractability_2021}, and the SHAP values can become less valuable for identifying structure-property relationships when using features that do not carry physical meaning. \cite{lundberg_unified_2017}

Saliency maps can provide an alternative method for understanding the contribution of features to a model's prediction. Saliency maps involve visualizing the gradient of each feature with respect to an output score, which can be intuitive for images where a saliency map will indicate pixel importance, but may not be useful when features are not inherently spatial.

\diff{

\section{Computational Performance Comparisons}

To further evaluate the performance of PCovC, we ran a number of PCA, LDA, PCovC, t-SNE, and UMAP fits on synthetic data. PCA and LDA fits were performed using \texttt{scikit-learn \cite{pedregosa_scikit-learn_2011}} , t-SNE fits were performed using \texttt{openTSNE} \cite{policar_opentsne_2024} with default hyperparameters, and UMAP fits were performed using \texttt{umap-learn} \cite{mcinnes_umap_2020} with default hyperparameters. All benchmarks were ran on a server with AMD EPYC 7763 processors with 10 GB of RAM.}

\begin{figure*}[ht!]
\begin{subfigure}[t]{0.5\linewidth}
    \includegraphics[width=\linewidth]{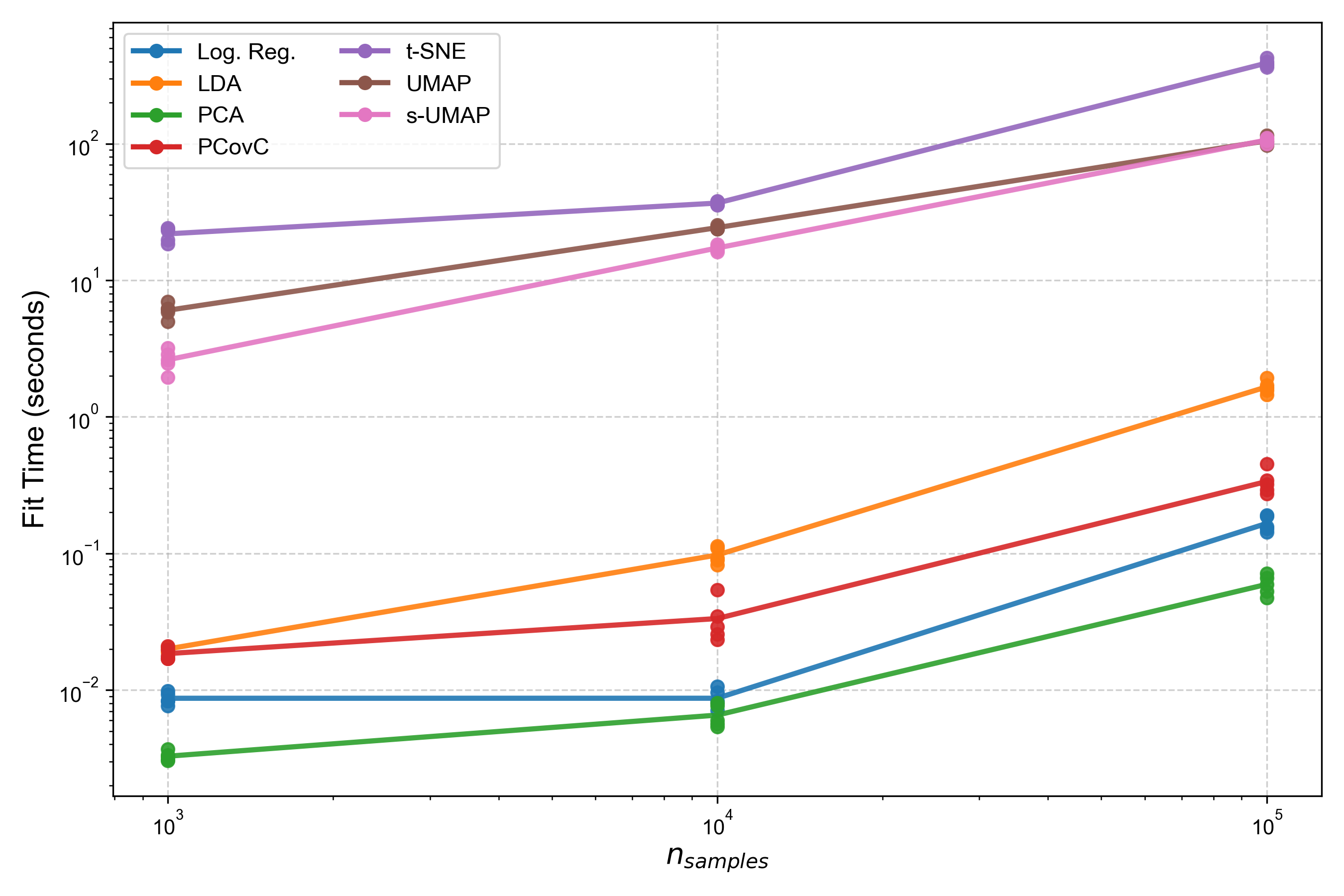}
    \caption{\diff{\textbf{Performance of PCovC, logistic regression, LDA, PCA, t-SNE, UMAP, and supervised UMAP with $n_{\mathbf{samples}}=1,000, 10,000, $ and $100,000$.} Here, we apply each method on a synthetic dataset with $n_{\mathbf{features}}=100$ and $n_{\mathbf{classes}}=2$.}}
\end{subfigure}\begin{subfigure}[t]{0.5\linewidth}
    \includegraphics[width=\linewidth]{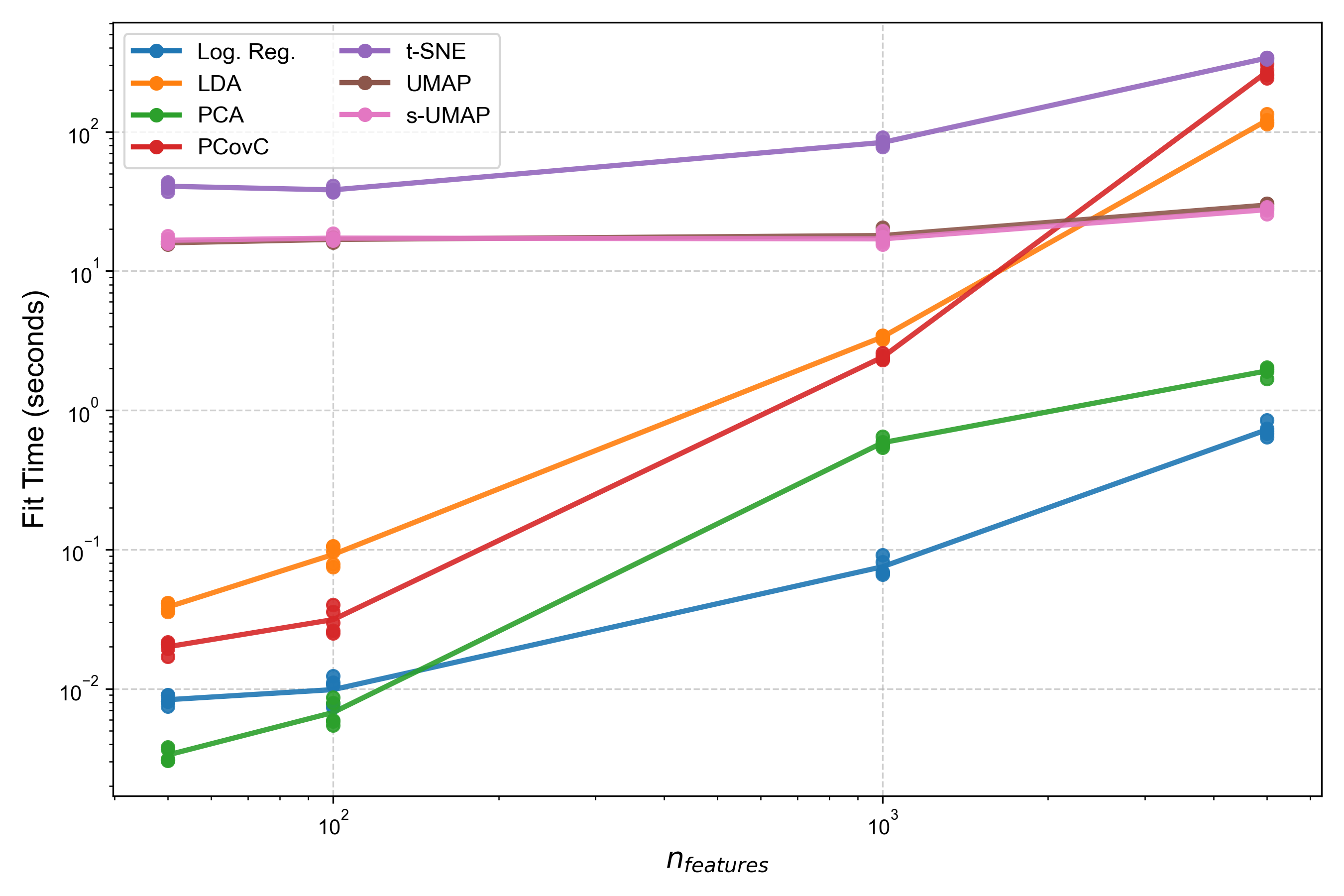}
    \caption{\diff{\textbf{Performance of PCovC, logistic regression, LDA, PCA, t-SNE, UMAP, and supervised UMAP with $n_{\mathbf{features}}=50, 100, 1,000, $ and $5,000$.} Here, we apply each method on a synthetic dataset with $n_{\mathbf{samples}}=10,000$ and $n_{\mathbf{classes}}=2$.}}
\end{subfigure}
\begin{subfigure}[t]{0.5\linewidth}
    \includegraphics[width=\linewidth]{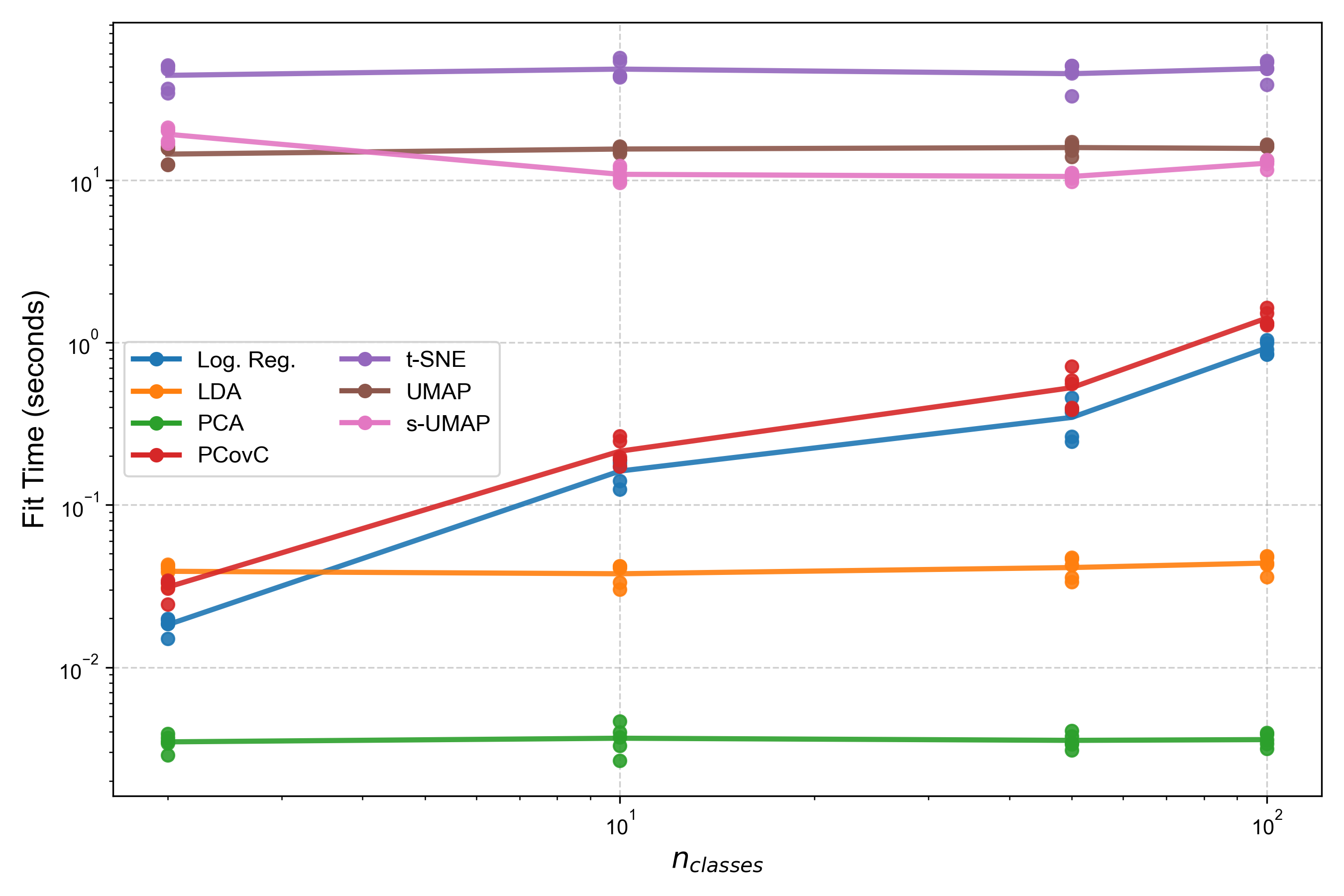}
    \caption{\diff{\textbf{Performance of PCovC, logistic regression, LDA, PCA, t-SNE, UMAP, and supervised UMAP with $n_{\mathbf{classes}}=2, 10, 50, $ and $100$.} Here, we apply each method on a synthetic dataset with $n_{\mathbf{samples}}=10,000$ and $n_{\mathbf{features}}=2$.}}
\end{subfigure}
\caption{\textbf{Performance benchmarks of PCovC with existing methods.}}
\end{figure*}

\diff{Generally, these results indicate that PCovC performs competitively with existing linear dimensionality reduction algorithms and can be faster than UMAP or t-SNE, as long as the data is not extremely high-dimensional.

\section{Comparison with Existing Methods}}

\diff{Here, we compare the maps obtained with PCovC to other popular data visualization techniques. Maps were made using the data discussed in Sec.~\ref{sec:tox}, with the ``tox21-ache-p3 (-1)" assay results as labels.} 

\begin{figure*}[ht!]
    \includegraphics[width=\linewidth]{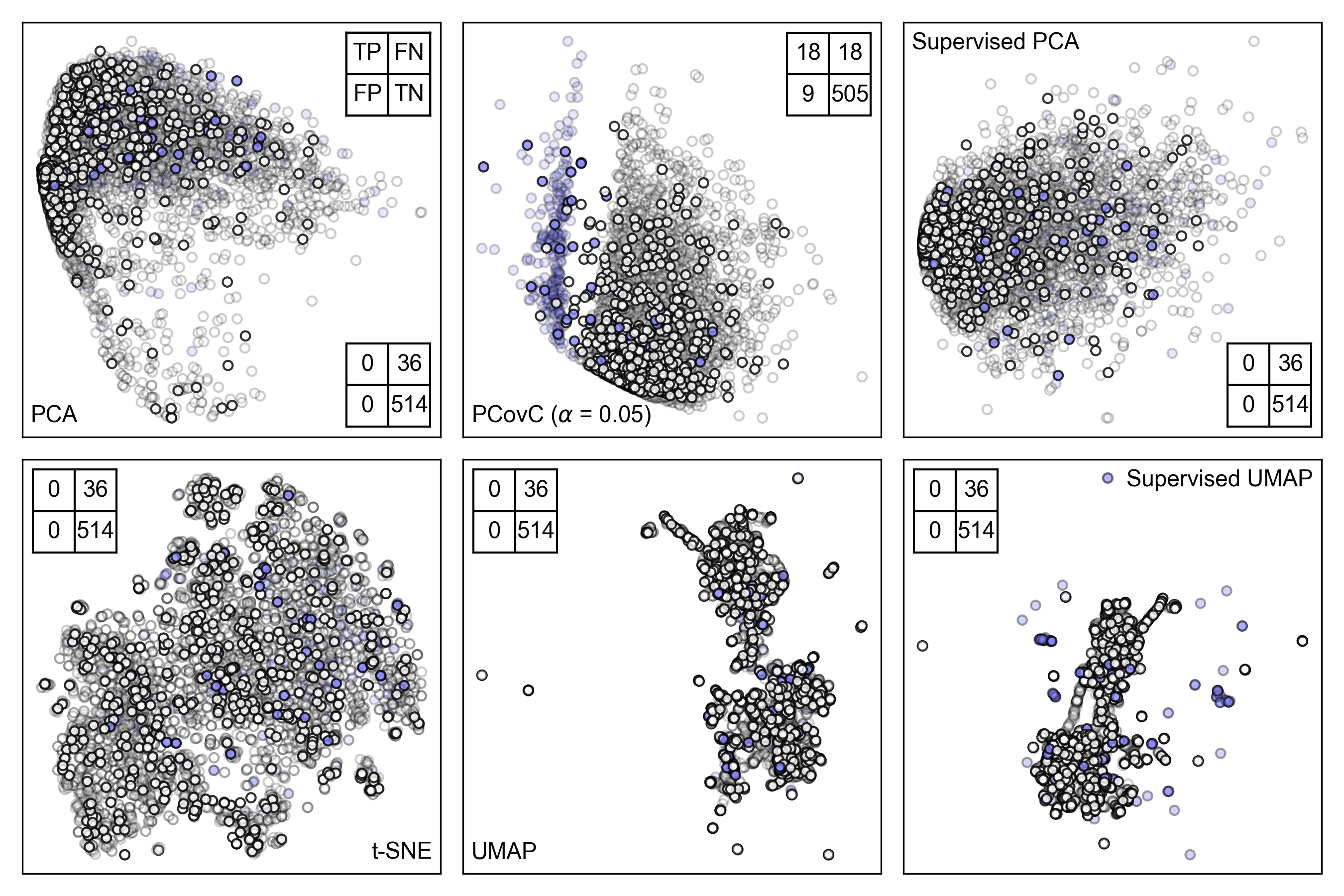}
    \caption{\diff{\textbf{Comparison of PCovC to PCA, supervised PCA, t-SNE, unsupervised UMAP, and supervised UMAP for neurotoxicity classification.} White and blue points indicate ``non-toxic'' and ``toxic'' molecules, respectively. Opacity denotes train (translucent)/ test (opaque) split. In each panel is a confusion matrix showing the accuracy of logistic regression on the testing data as they appear in the resulting map, where ``TN'' indicates the number of ``true negatives'', ``FP'' indicates the number of ``false positives'', and so on.}}
\end{figure*}

\end{document}

\title{Interpretable Visualizations of Data Spaces for Classification Problems: \\Supplementary Information}

\author{Christian Jorgensen}
\affiliation{Department of Chemical and Biological Engineering, University of Wisconsin - Madison, Madison, WI, USA}
\author{Arthur Y. Lin}
\affiliation{Department of Chemical and Biological Engineering, University of Wisconsin - Madison, Madison, WI, USA}
\author{Rhushil Vasavada}
\affiliation{Department of Computer Sciences, University of Wisconsin - Madison, Madison, WI, USA}
\author{Rose K. Cersonsky}
\email{rose.cersonsky@wisc.edu}
\affiliation{Department of Chemical and Biological Engineering, University of Wisconsin - Madison, Madison, WI, USA}
\affiliation{Department of Materials Science and Engineering, University of Wisconsin - Madison, Madison, WI, USA}
\affiliation{Data Science Institute, University of Wisconsin - Madison, Madison, WI, USA}

\maketitle

\section{Additional Material Included in this SI}
Included in the SI are two types of files for better understanding the PCovC framework. The \texttt{mp4} files animate the visualizations included in the main text as a function of $\alpha$, smoothly showing the transformation of the latent-space projection as we modulate the loss function. The \texttt{json} files are formatted for the \texttt{chemiscope} visualization tool available at \url{chemiscope.org}. Best effort has been made to convert chemical identifiers in all underlying datasets to visualizable structures for the \texttt{3dmol} viewer.

\section{Additional Figures}

\clearpage

\subsection{What makes a molecule toxic? Exposing decision boundaries in toxicity classifications}

\begin{figure*}[ht!]
    \includegraphics[width=\linewidth]{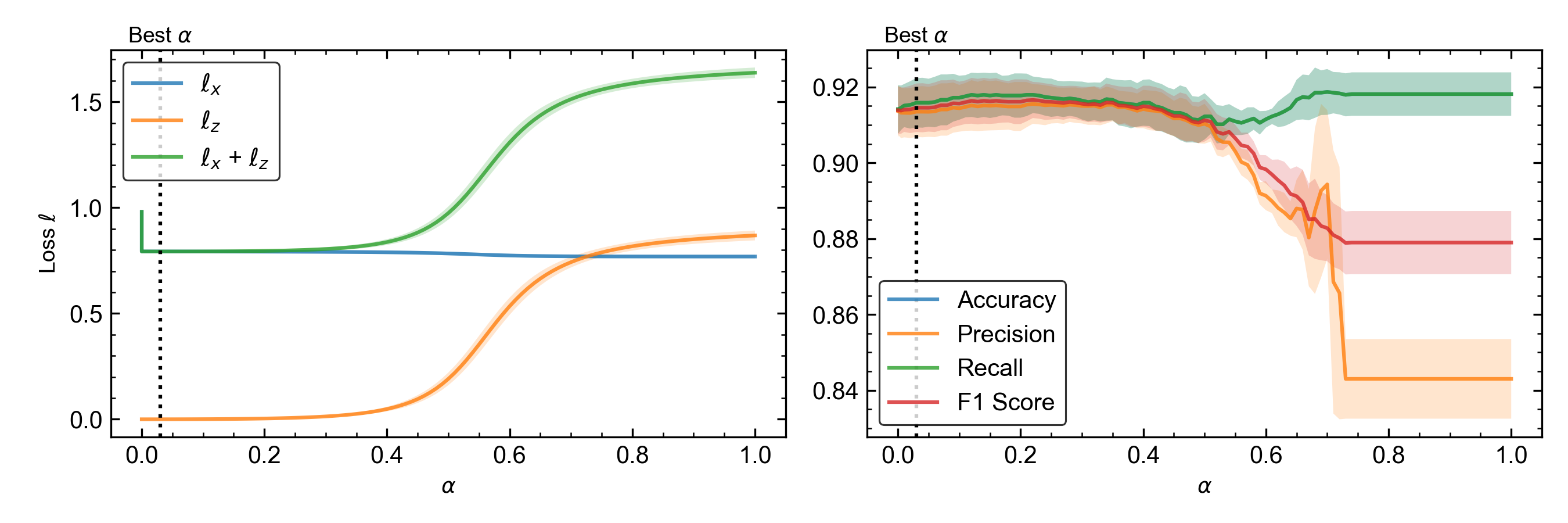}
    \caption{\textbf{5-fold cross-validation results for the ``tox21-ache-p1 (-1)'' assay with the multilabel dataset.} The left plot shows the unsupervised, supervised, and summed PCovC loss terms with varying $\alpha$ values on the validation set, and the right plot shows the validation accuracy, precision, recall, and F1 score of a logistic regression model learned in the latent space with varying $\alpha$ values. Here, $\alpha=0.03$ optimizes the PCovC loss.}
\end{figure*}

\begin{figure*}[ht!]
    \includegraphics[width=\linewidth]{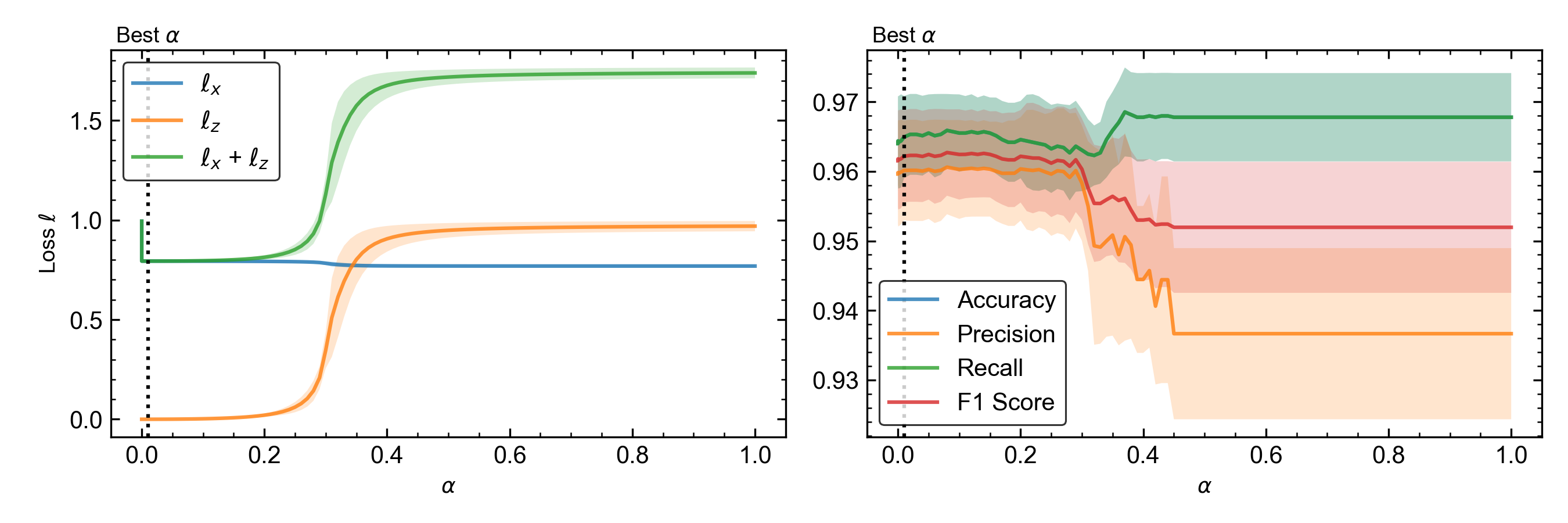}
    \caption{\textbf{5-fold cross-validation results for the ``tox21-ache-p3 (-1)'' assay with the multilabel dataset.} The left plot shows the unsupervised, supervised, and summed PCovC loss terms with varying $\alpha$ values on the validation set, and the right plot shows the validation accuracy, precision, recall, and F1 score of a logistic regression model learned in the latent space with varying $\alpha$ values. Here, $\alpha=0.01$ optimizes the PCovC loss.}
\end{figure*}

\begin{figure*}[ht!]
    \includegraphics[width=\linewidth]{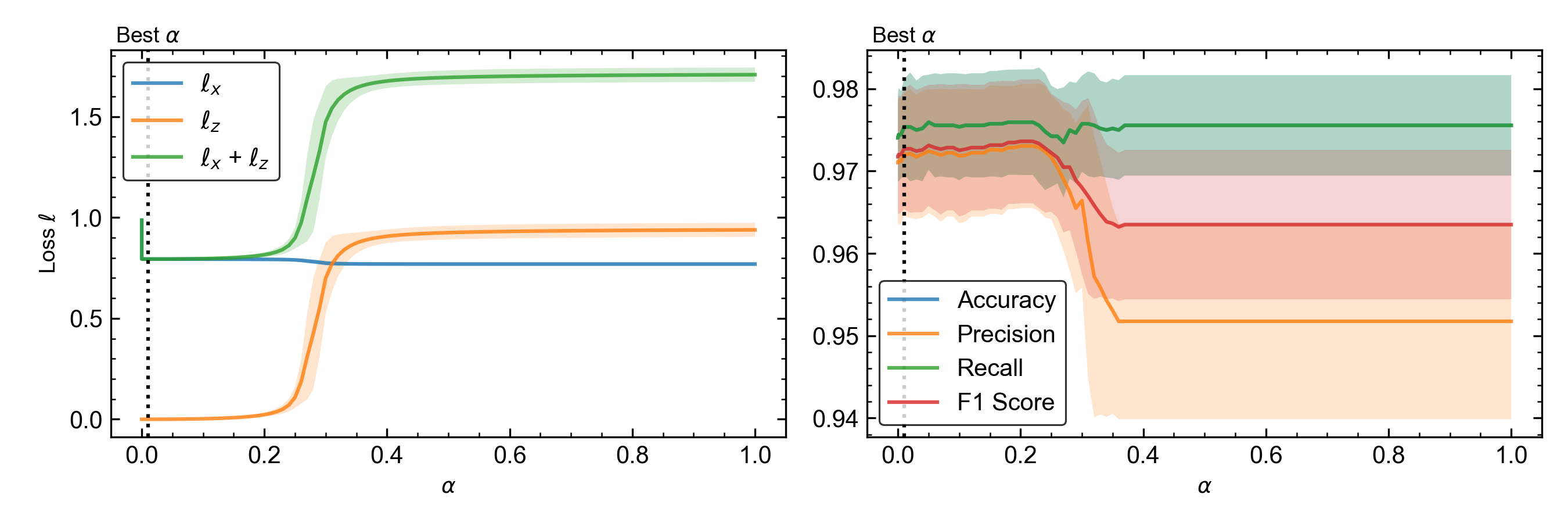}
\caption{\textbf{5-fold cross-validation results for the ``tox21-ache-p5 (-1)'' assay with the multilabel dataset.} The left plot shows the unsupervised, supervised, and summed PCovC loss terms with varying $\alpha$ values on the validation set, and the right plot shows the validation accuracy, precision, recall, and F1 score of a logistic regression model learned in the latent space with varying $\alpha$ values. Here, $\alpha=0.01$ optimizes the PCovC loss.}
\end{figure*}

\begin{figure*}[ht!]
    \includegraphics[width=\linewidth]{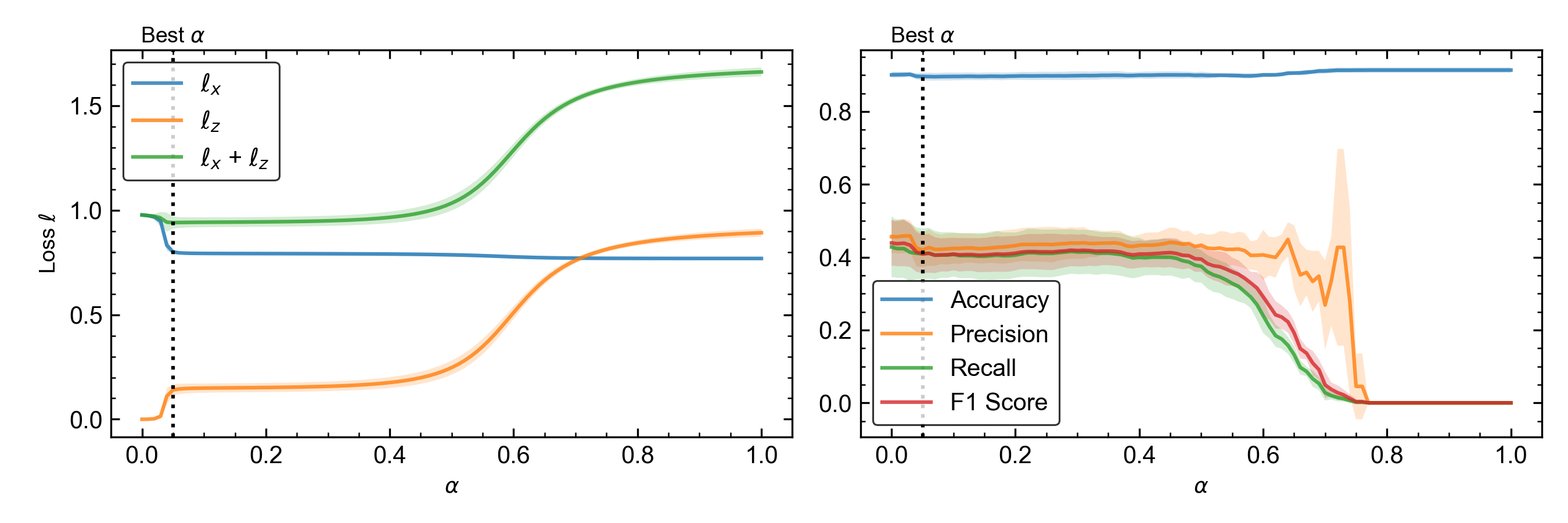}
    \caption{\textbf{5-fold cross-validation results for the multilabel dataset with the ``tox21-ache-p1 (-1)" and ``tox21-ache-p3 (-1)" assays used for training.} The left plot shows the unsupervised, supervised, and summed PCovC loss terms with varying $\alpha$ values on the validation set, and the right plot shows the validation accuracy, precision, recall, and F1 score of a logistic regression model learned in the latent space with varying $\alpha$ values. Precisions, recalls, and F1 scores were each weighted by the number of toxic compounds by each label to account for label inbalance. Here, $\alpha=0.05$ optimizes the PCovC loss.}
\end{figure*}

\begin{figure*}[ht!]
    \includegraphics[width=\linewidth]{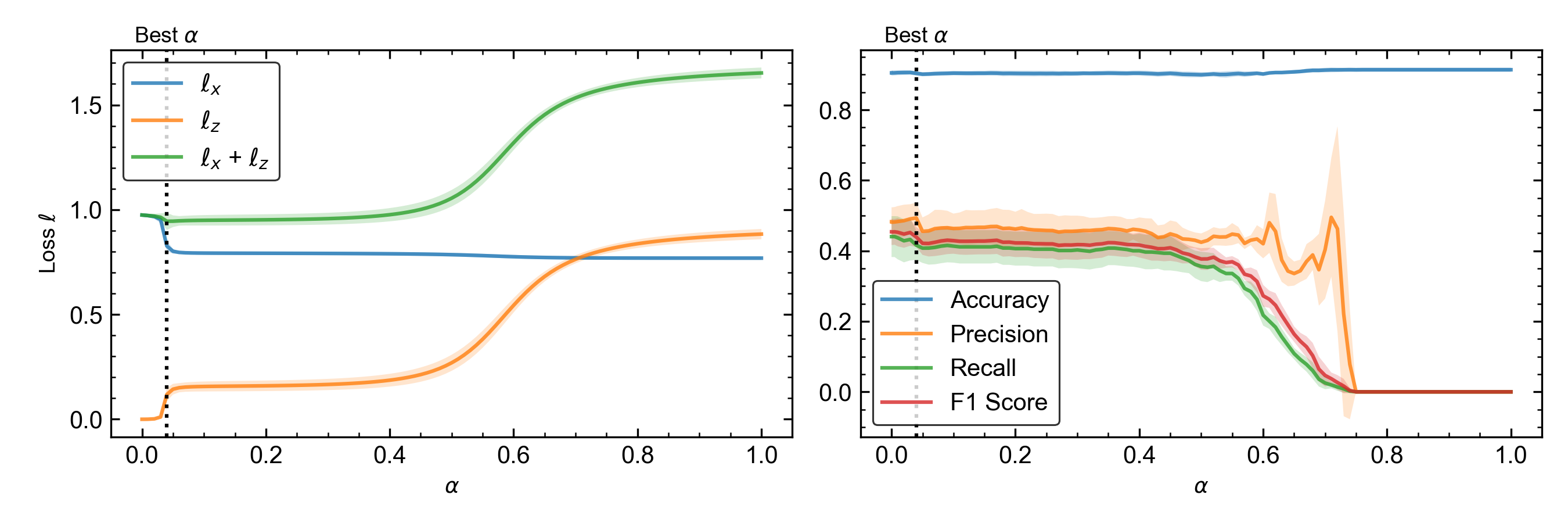}
     \caption{\textbf{5-fold cross-validation results for the multilabel dataset with the ``tox21-ache-p1 (-1)" and ``tox21-ache-p5 (-1)" assays used for training.} The left plot shows the unsupervised, supervised, and summed PCovC loss terms with varying $\alpha$ values on the validation set, and the right plot shows the validation accuracy, precision, recall, and F1 score of a logistic regression model learned in the latent space with varying $\alpha$ values. Precisions, recalls, and F1 scores were each weighted by the number of toxic compounds by each label to account for label inbalance. Here, $\alpha=0.04$ optimizes the PCovC loss.}
\end{figure*}

\begin{figure*}[ht!]
    \includegraphics[width=\linewidth]{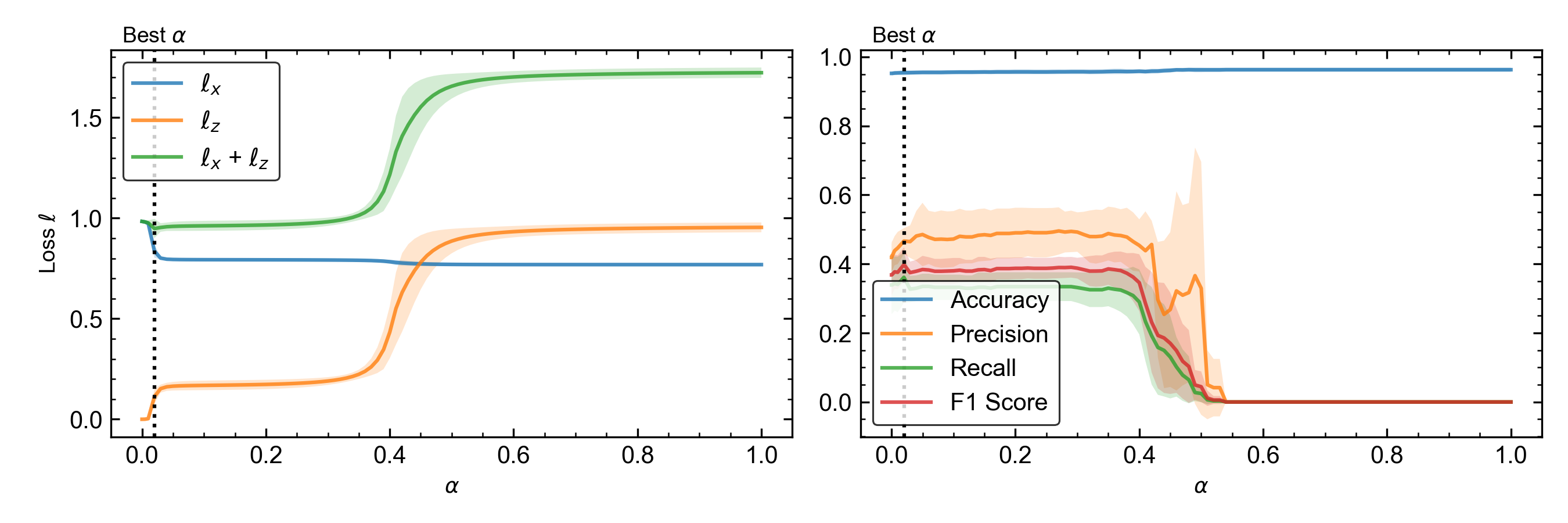}
     \caption{\textbf{5-fold cross-validation results for the multilabel dataset with the ``tox21-ache-p3 (-1)" and ``tox21-ache-p5 (-1)" assays used for training.} The left plot shows the unsupervised, supervised, and summed PCovC loss terms with varying $\alpha$ values on the validation set, and the right plot shows the validation accuracy, precision, recall, and F1 score of a logistic regression model learned in the latent space with varying $\alpha$ values. Precisions, recalls, and F1 scores were each weighted by the number of toxic compounds by each label to account for label inbalance. Here, $\alpha=0.02$ optimizes the PCovC loss.}
\end{figure*}

\begin{figure*}[ht!]
    \includegraphics[width=\linewidth]{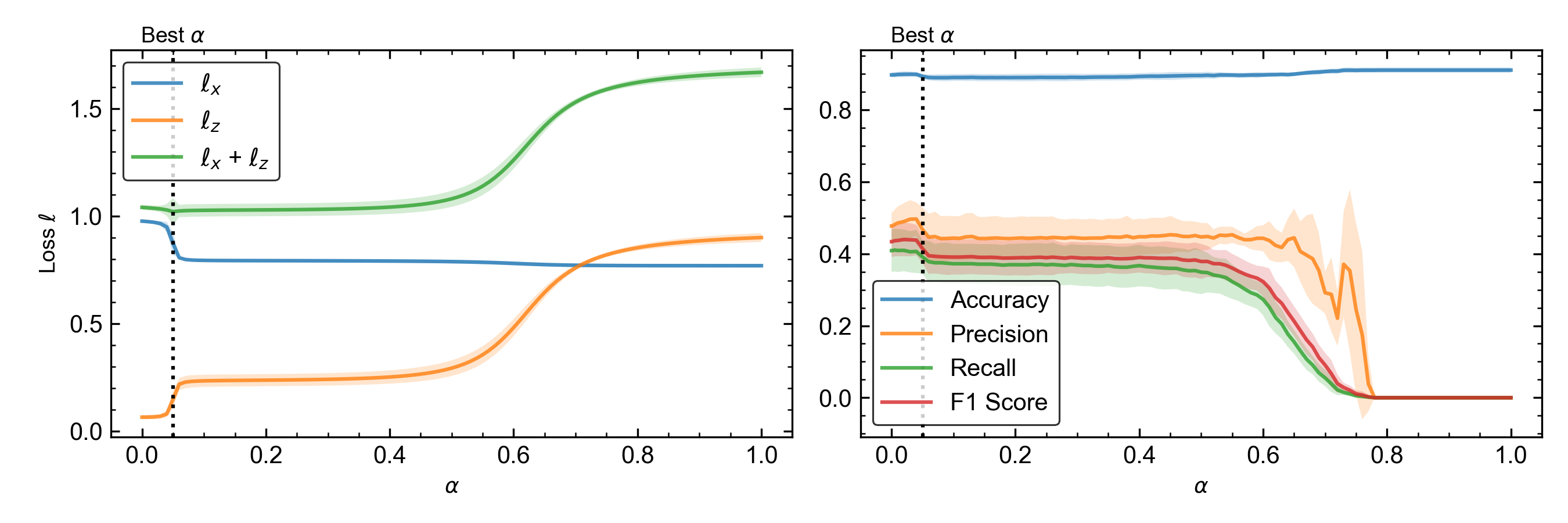}
     \caption{\textbf{5-fold cross-validation results for the multilabel dataset with the ``tox21-ache-p1 (-1)", ``tox21-ache-p3 (-1)", and ``tox21-ache-p5 (-1)" assays used for training.} The left plot shows the unsupervised, supervised, and summed PCovC loss terms with varying $\alpha$ values on the validation set, and the right plot shows the validation accuracy, precision, recall, and F1 score of a logistic regression model learned in the latent space with varying $\alpha$ values. Precisions, recalls, and F1 scores were each weighted by the number of toxic compounds by each label to account for label inbalance. Here, $\alpha=0.05$ optimizes the PCovC loss.}
\end{figure*}

\begin{figure}[ht!]
    \includegraphics[width=\linewidth]{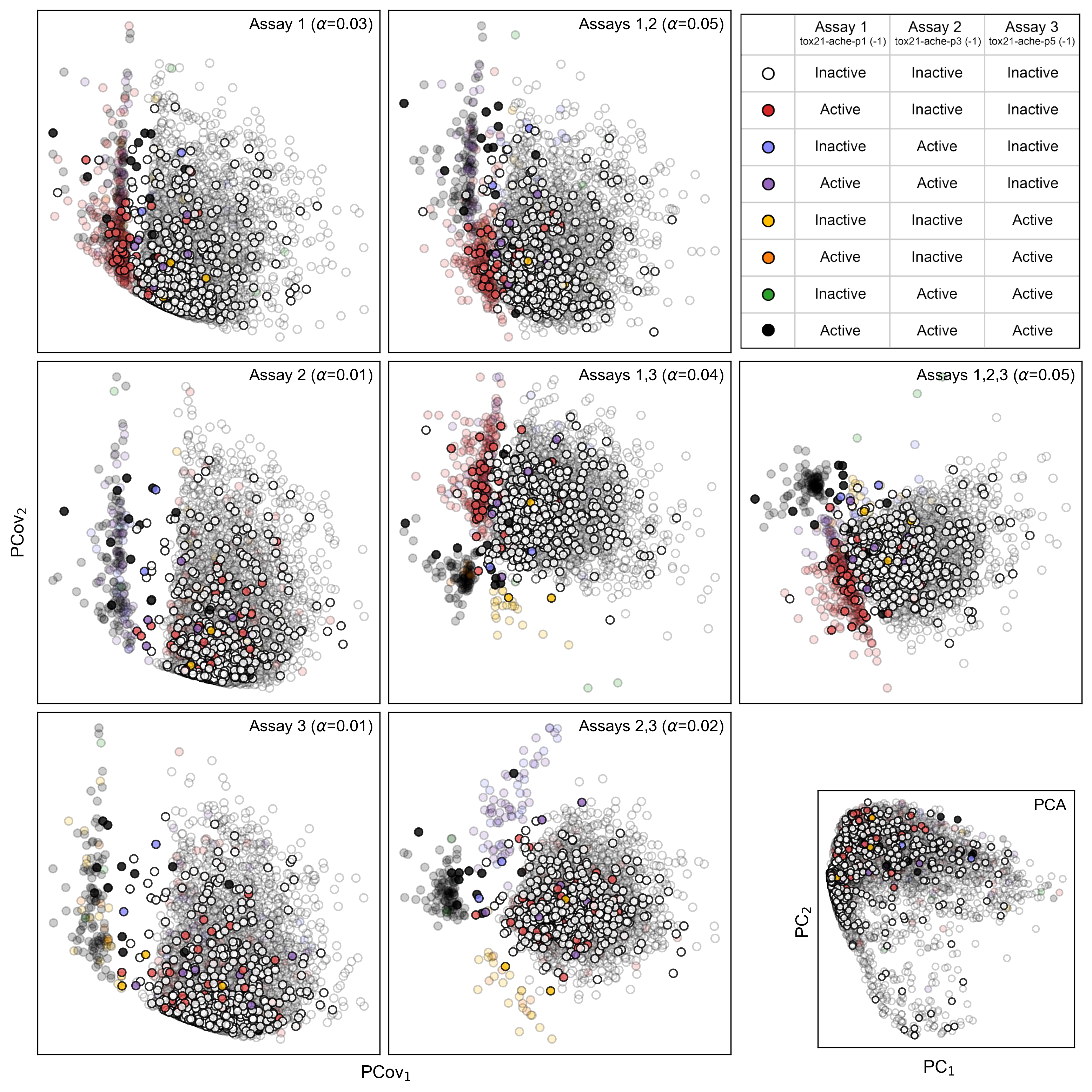}
    \caption{\textbf{Full set of multilabel PCovC maps considering multiple Tox21 neurotoxicity assays} for classification, (left) three single-assay maps, (center) three double-assay maps, (right) the triple-assay map, and the PCA projection (lower right). Color corresponds to the values for the three assays, shown in the table in the top right. Opacity corresponds to the train/test split.
    }
\end{figure}

\clearpage

\subsection{Parsing multi-class datasets: spectral analysis of organosulfur bonding environments}

\begin{figure}[ht!]
    \includegraphics[width=\linewidth]{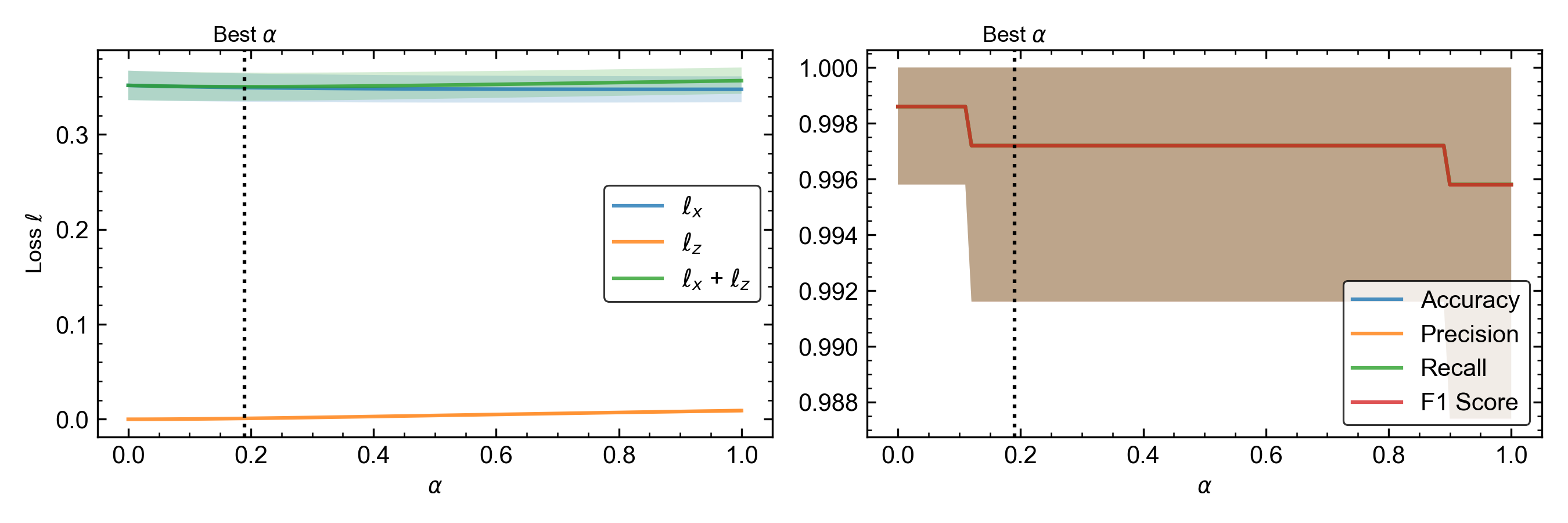}
    \caption{\textbf{5-fold cross-validation results for oxidation state mapping.} The left plot shows the unsupervised, supervised, and summed PCovC loss terms with varying $\alpha$ values on the validation set, and the right plot shows the validation accuracy, precision, recall, and F1 score of a logistic regression model learned in the latent space with varying $\alpha$ values. Precisions, recalls, and F1 scores were each weighted by the number of compounds in each oxidation state to account for class inbalance, and are all overlapping. Here, $\alpha=0.19$ minimizes the PCovC loss.}
\end{figure}

\begin{figure}[ht!]
    \includegraphics[width=\linewidth]{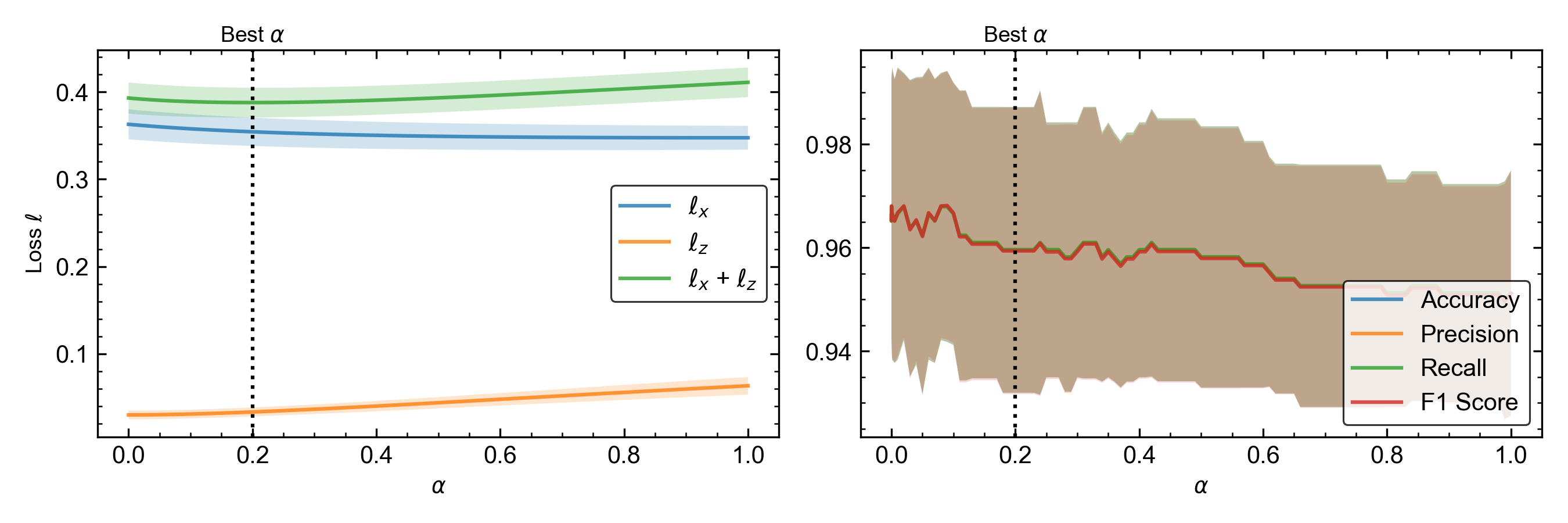}
    \caption{\textbf{5-fold cross-validation results for bond type mapping.} The left plot shows the unsupervised, supervised, and summed PCovC loss terms with varying $\alpha$ values on the validation set, and the right plot shows the validation accuracy, precision, recall, and F1 score of a logistic regression model learned in the latent space with varying $\alpha$ values. Precisions, recalls, and F1 scores were each weighted by the number of compounds in each bond type to account for class inbalance, and are all overlapping. Here, $\alpha=0.20$ minimizes the PCovC loss.}
\end{figure}

\begin{figure}[ht!]
    \includegraphics[width=\linewidth]{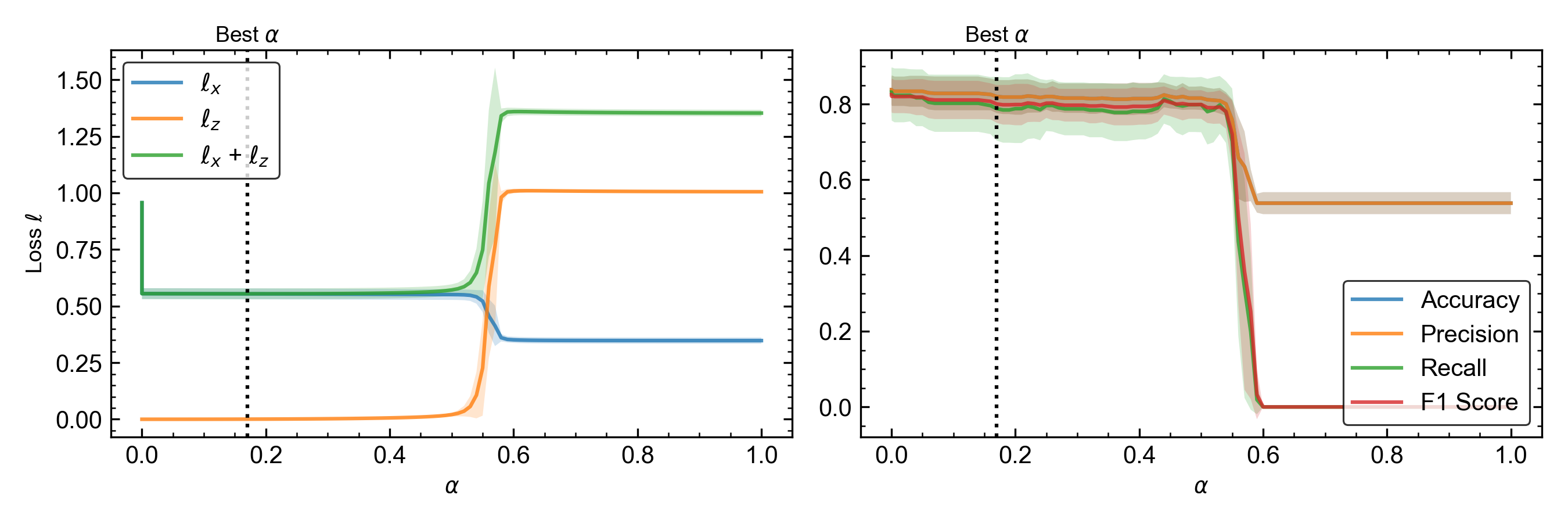}
    \caption{\textbf{5-fold cross-validation results for oxidation state mapping.} The left plot shows the unsupervised, supervised, and summed PCovC loss terms with varying $\alpha$ values on the validation set, and the right plot shows the validation accuracy, precision, recall, and F1 score of a logistic regression model learned in the latent space with varying $\alpha$ values. Here, $\alpha=0.17$ minimizes the PCovC loss.}
\end{figure}

\begin{figure}[ht!]
    \includegraphics[width=\linewidth]{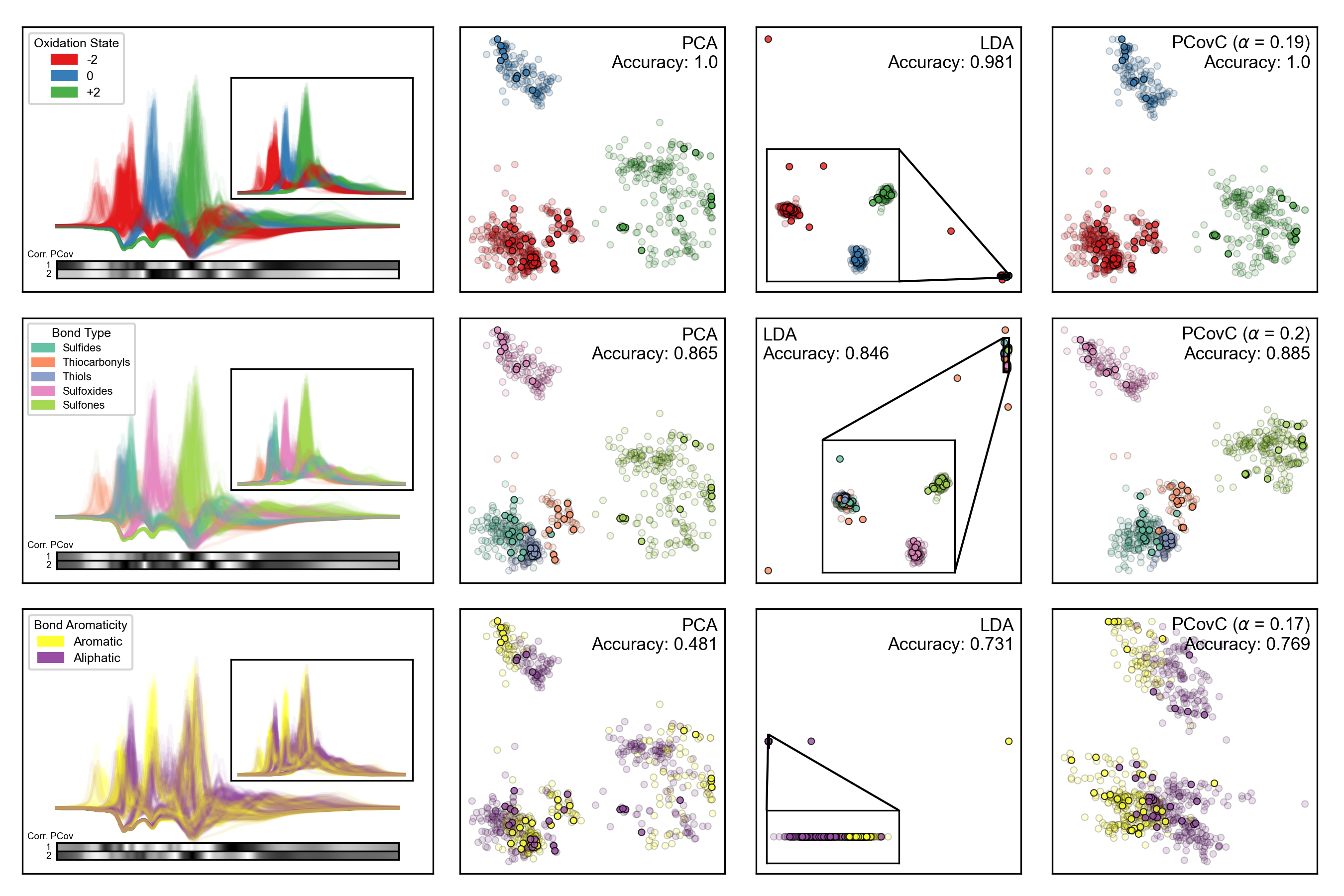}
    \caption{\textbf{Analysis of organosulfur dataset for different classification targets.} Rows correspond to classification based on oxidation state (top), bond type (center) and aromaticity (bottom). (left) Scaled XANES spectra, with raw spectra in the inset. Heatmaps underneath the spectra correspond to correlations with the two principal covariates. (right) PCA, LDA and PCovC projections, as shown in the main text.}
    \end{figure}

\clearpage
\subsection{Understanding feature importance in non-linear classification of metals/nonmetals}

{\begin{figure}[ht!]
    \includegraphics[width=\linewidth]{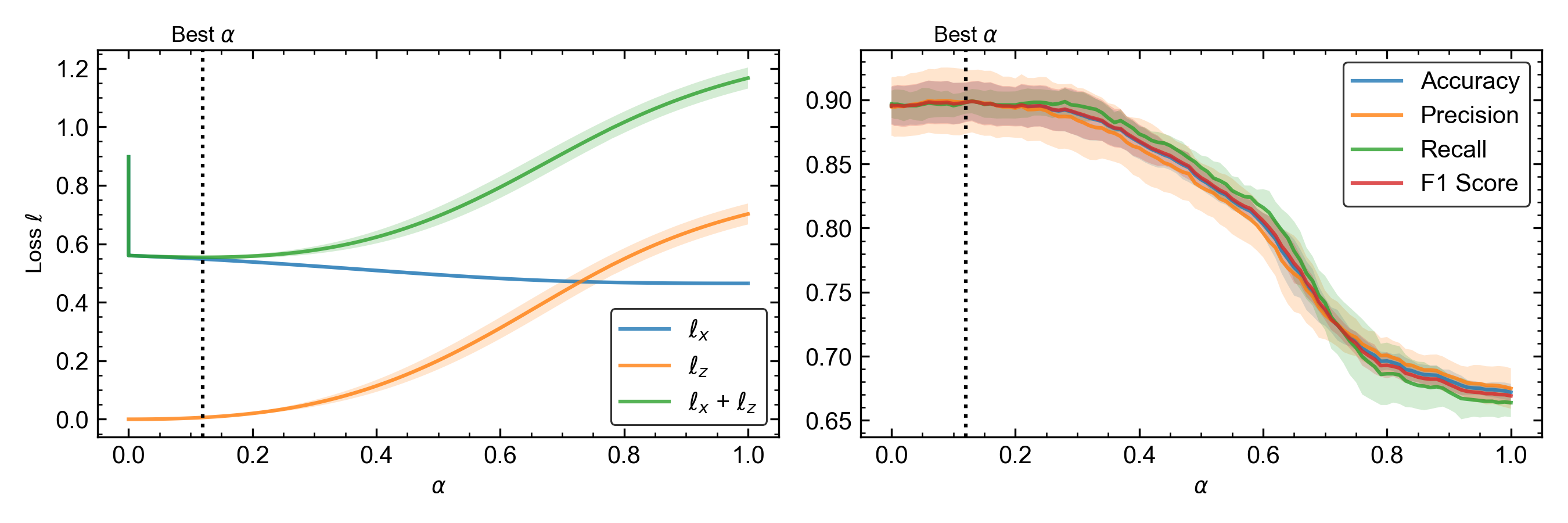}
    \caption{\textbf{5-fold cross-validation results for the metal/nonmetal dataset with KPCovC.} The left plot shows the unsupervised, supervised, and summed KPCovC loss terms with varying $\alpha$ values on the validation set, and the right plot shows the validation accuracy, precision, recall, and F1 score of a linear SVC model learned in the latent space with varying $\alpha$ values. Here, $\alpha=0.12$ minimizes the PCovC loss.}
    \end{figure}

\begin{figure}[ht!]
    \includegraphics[width=\linewidth]{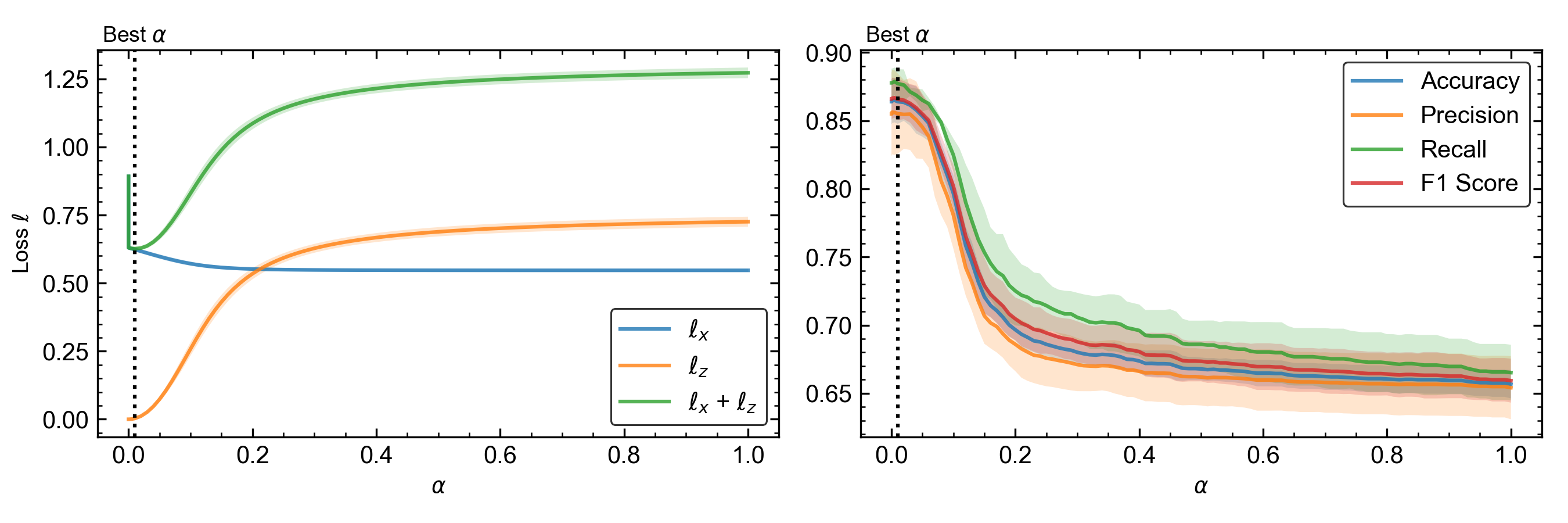}
     \caption{\textbf{5-fold cross-validation results for the metal/nonmetal dataset with PCovC.} The left plot shows the unsupervised, supervised, and summed PCovC loss terms with varying $\alpha$ values on the validation set, and the right plot shows the validation accuracy, precision, recall, and F1 score of a linear SVC model learned in the latent space with varying $\alpha$ values. Here, $\alpha=0.01$ minimizes the PCovC loss.}
    \end{figure}

\begin{figure}[ht!]
    \includegraphics[width=\linewidth, trim={1.5cm, 1.3cm, 1.5cm, 0.6cm}, clip]{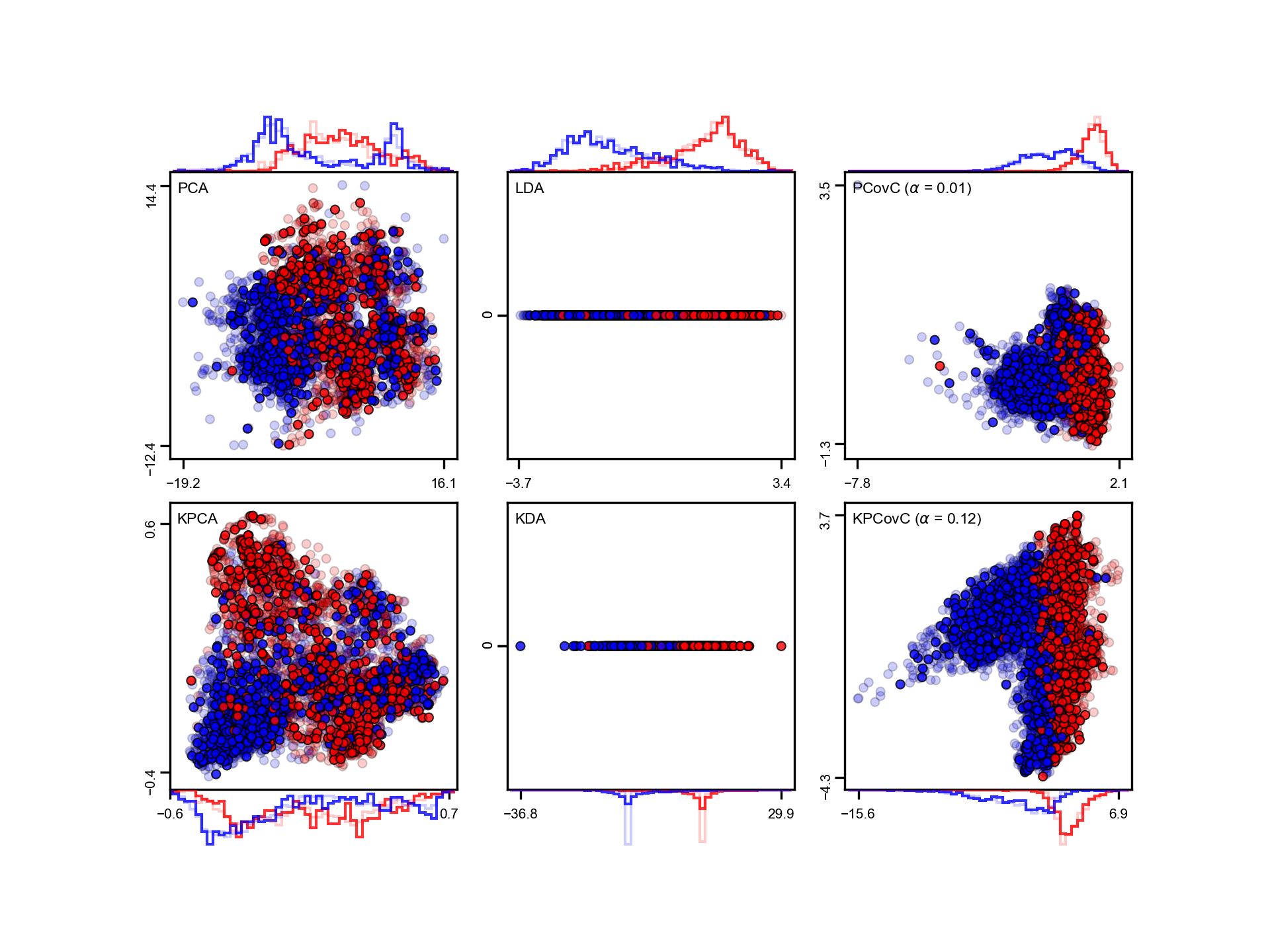}
    \caption{The classification of metal versus non-metal compounds requires a non-linear classification approach; here we recreate the figure from the main text also demonstrating an analogous linear approach (top) which yields similar conclusions.}
\end{figure}

\clearpage
We include below the correlation analyses of the low-dimensional maps and all 136 input features. Color indicates the absolute value of the Pearson correlation coefficient, and all labels are taken directly from the original dataset.

\begin{figure}[ht!]
    \includegraphics[width=0.95\linewidth]{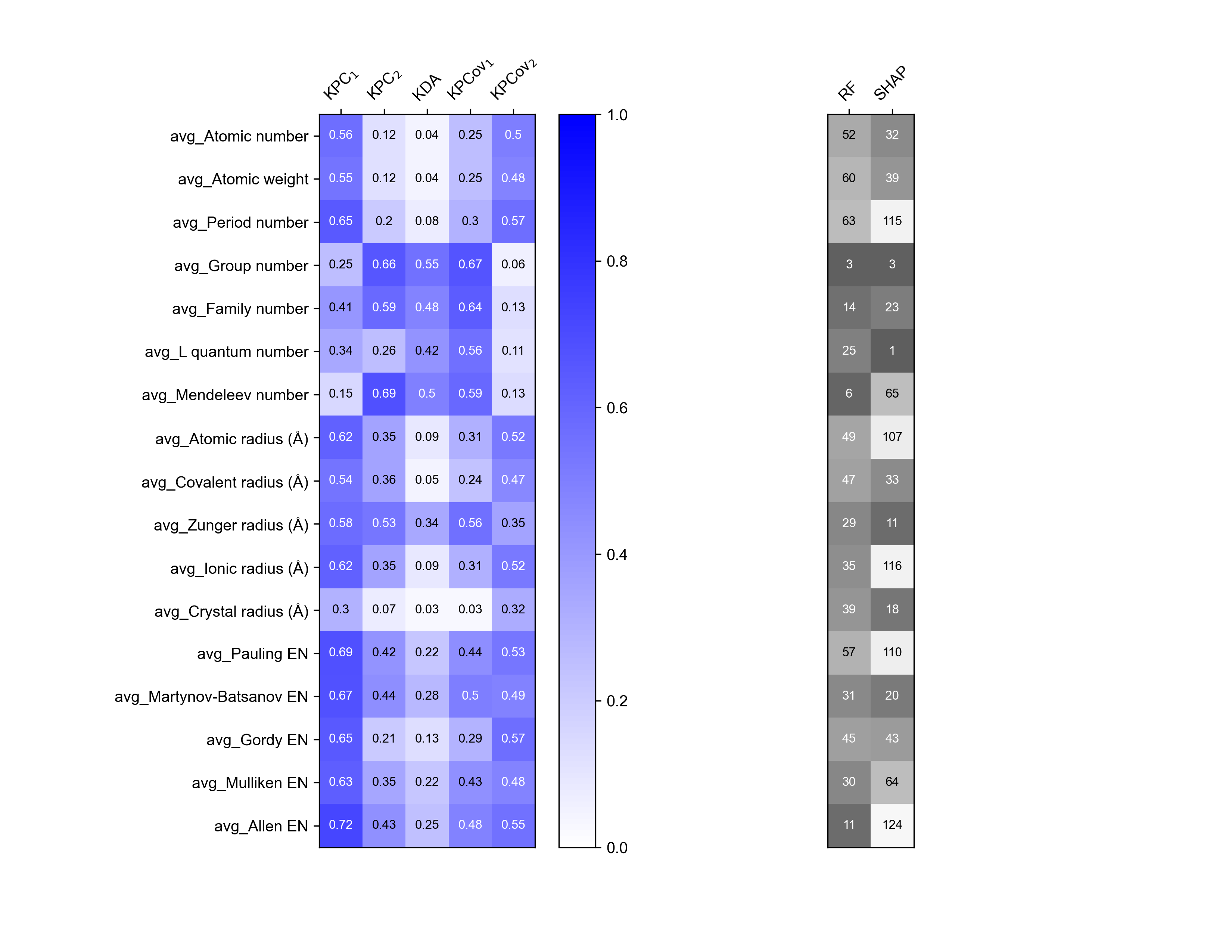}
    
    \includegraphics[width=0.95\linewidth]{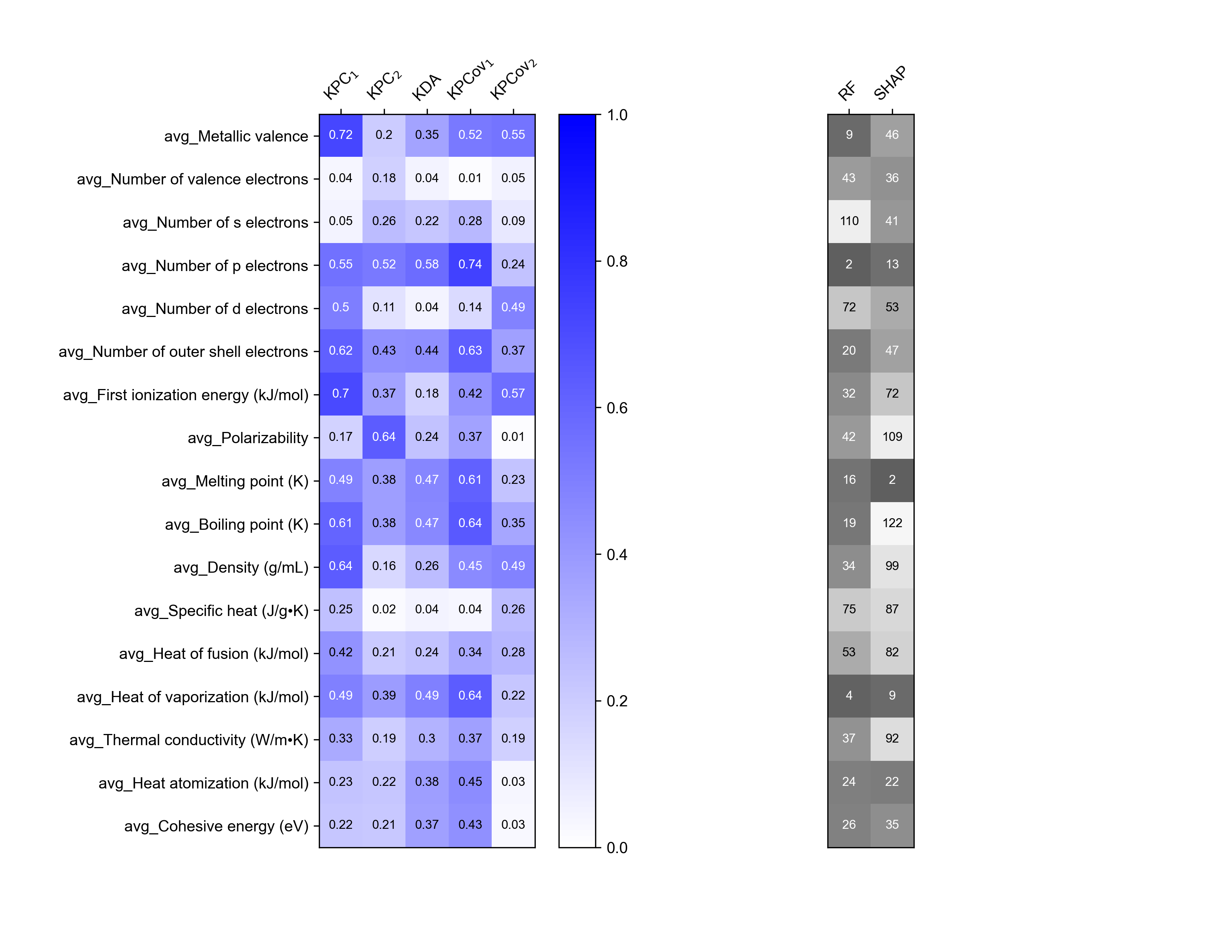}
\end{figure}

\begin{figure}[ht!]
    \includegraphics[width=0.95\linewidth]{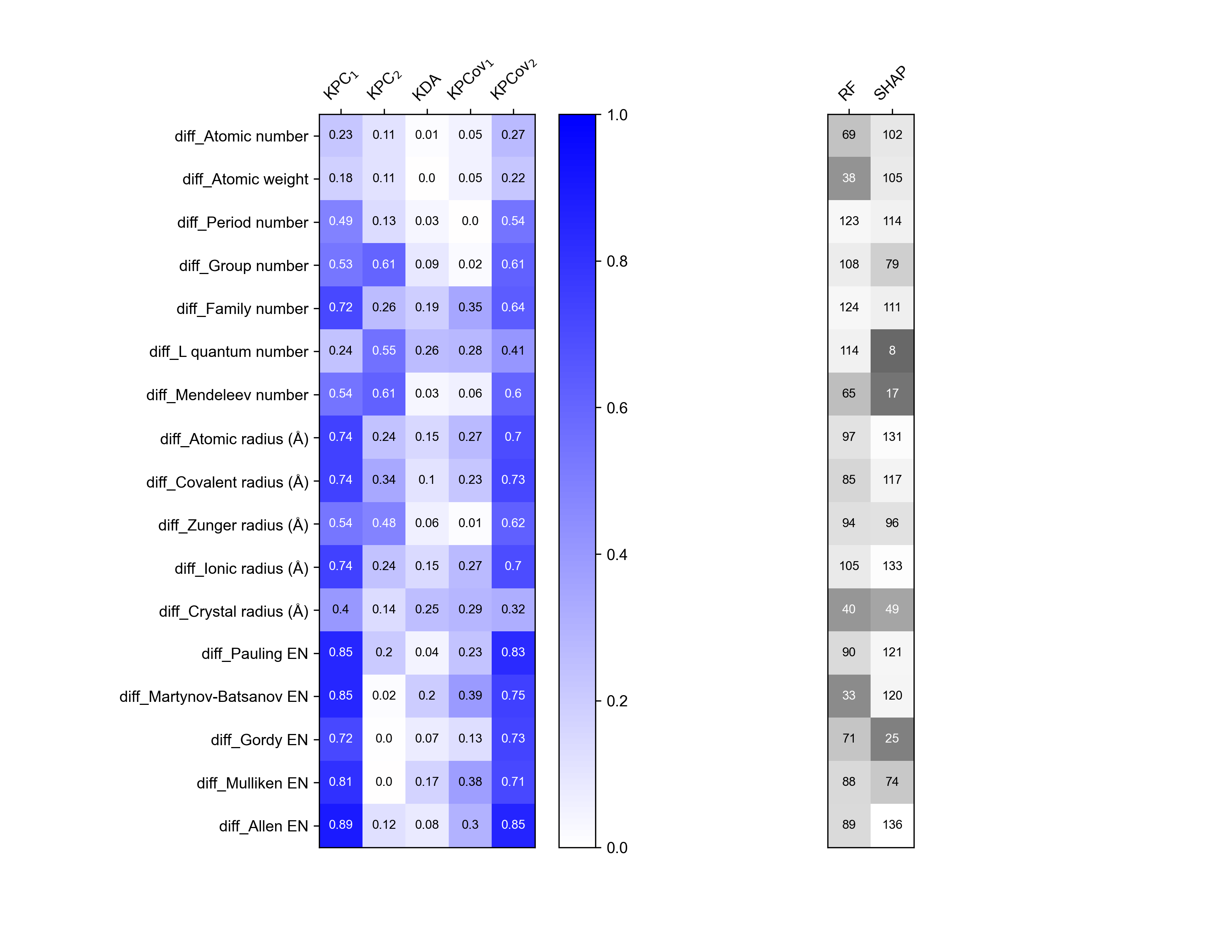}
    
    \includegraphics[width=0.95\linewidth]{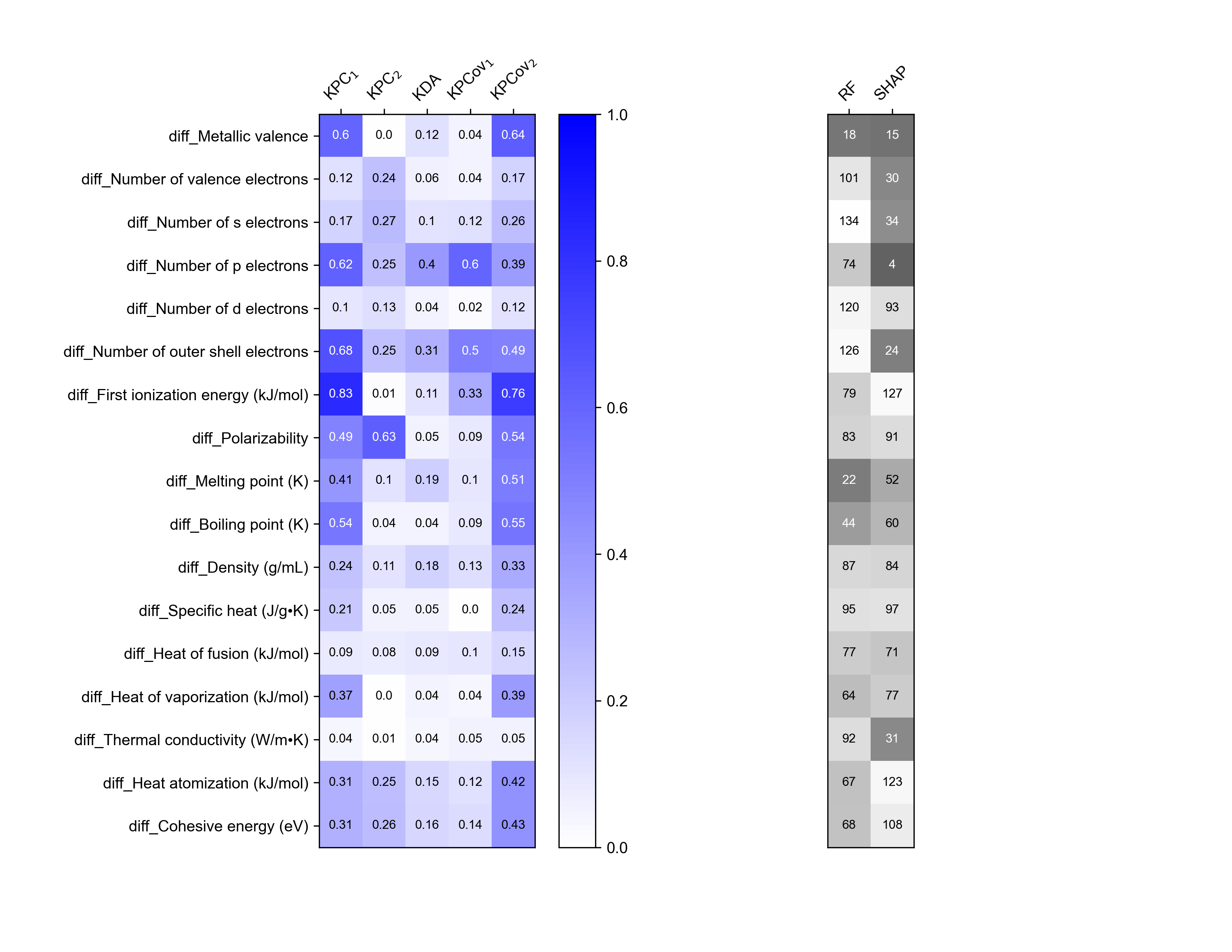}
\end{figure}

\begin{figure}[ht!]
    \includegraphics[width=0.95\linewidth]{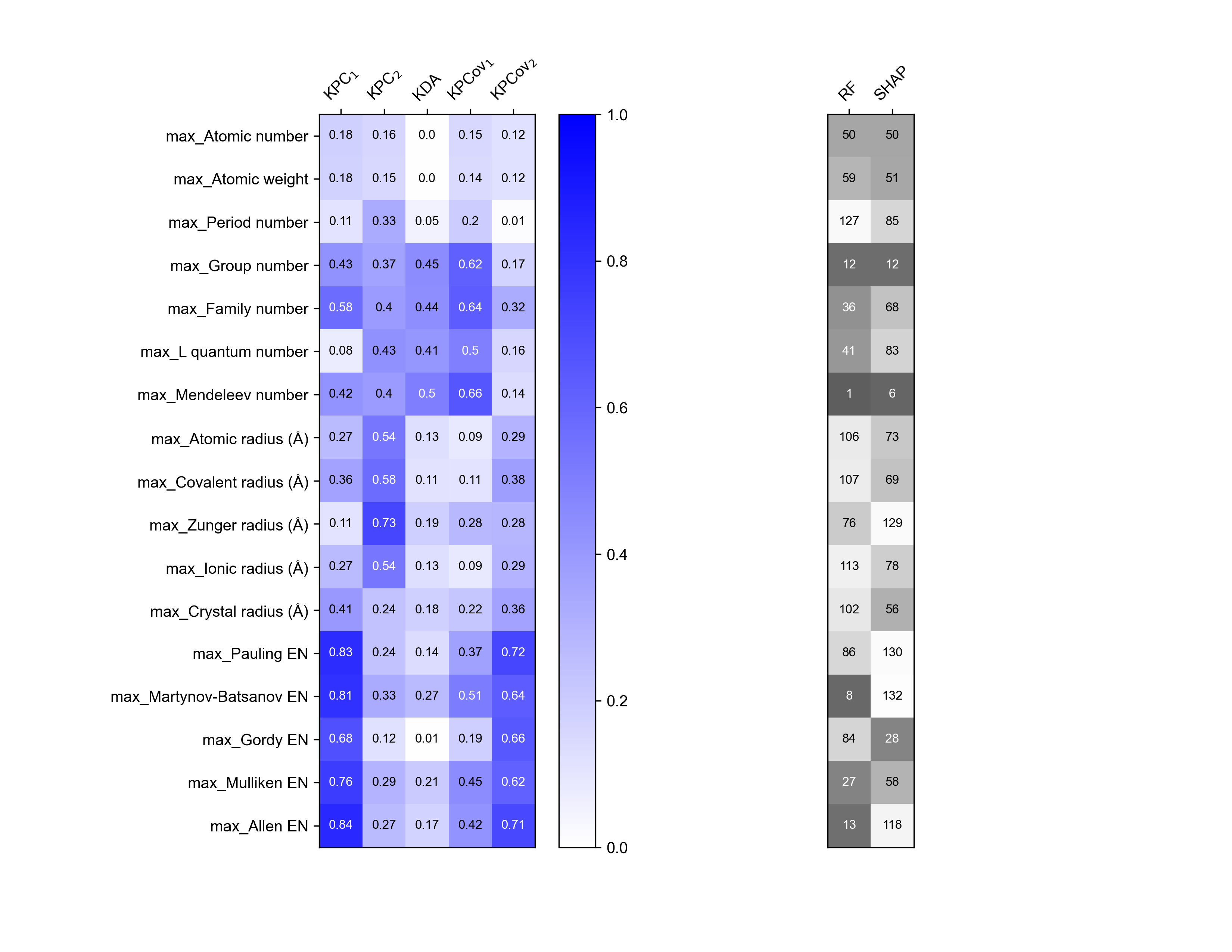}
    
    \includegraphics[width=0.95\linewidth]{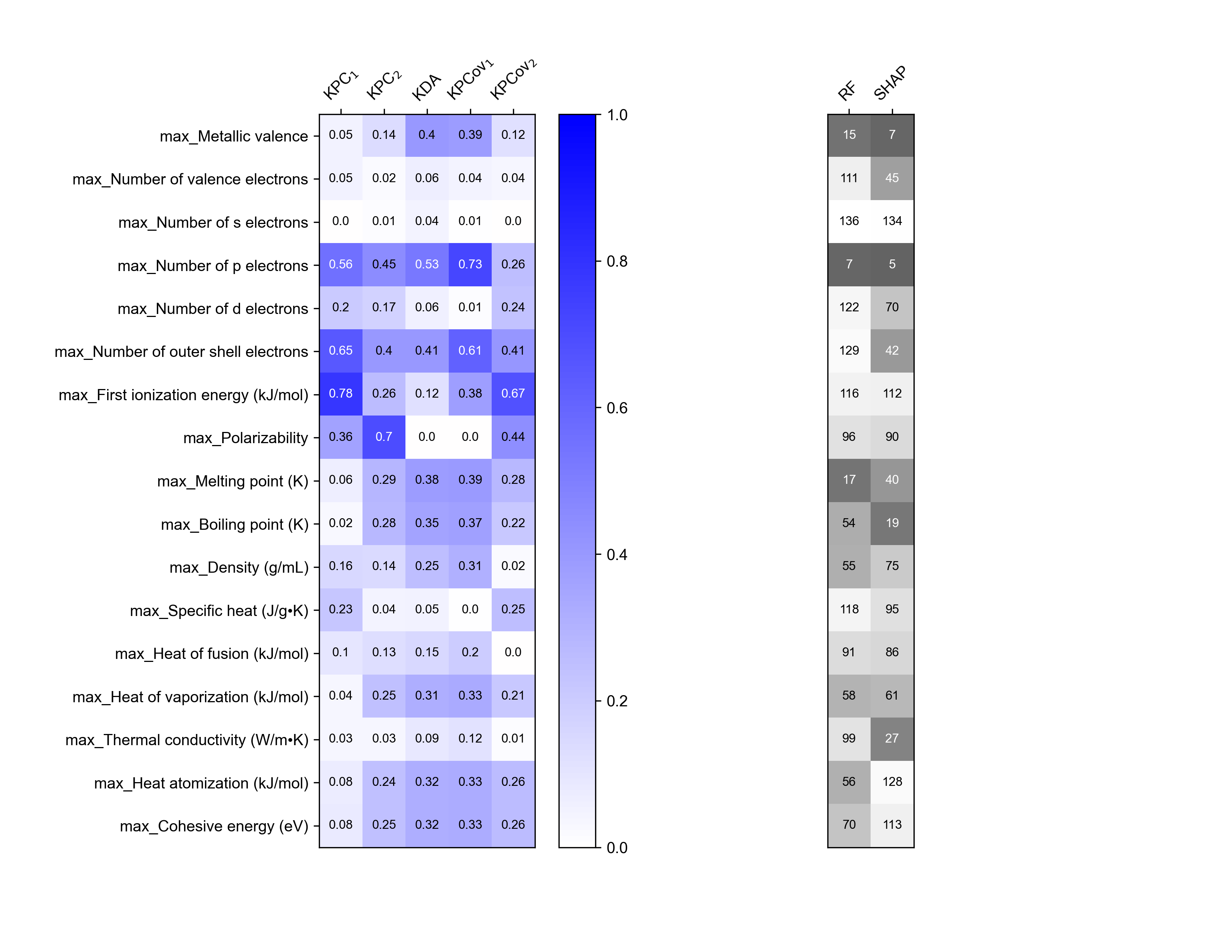}
\end{figure}

\begin{figure}[ht!]
    \includegraphics[width=0.95\linewidth]{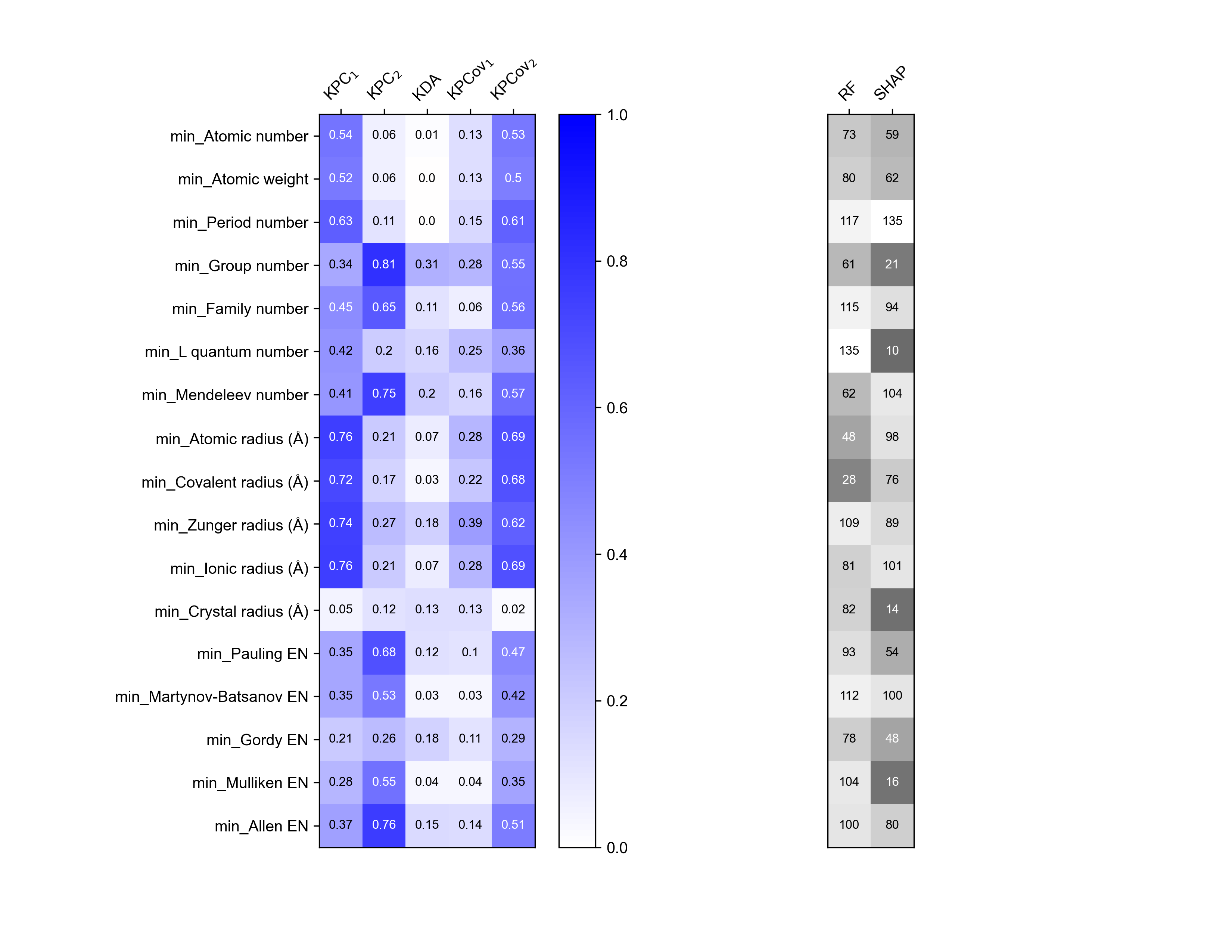}
    
    \includegraphics[width=0.95\linewidth]{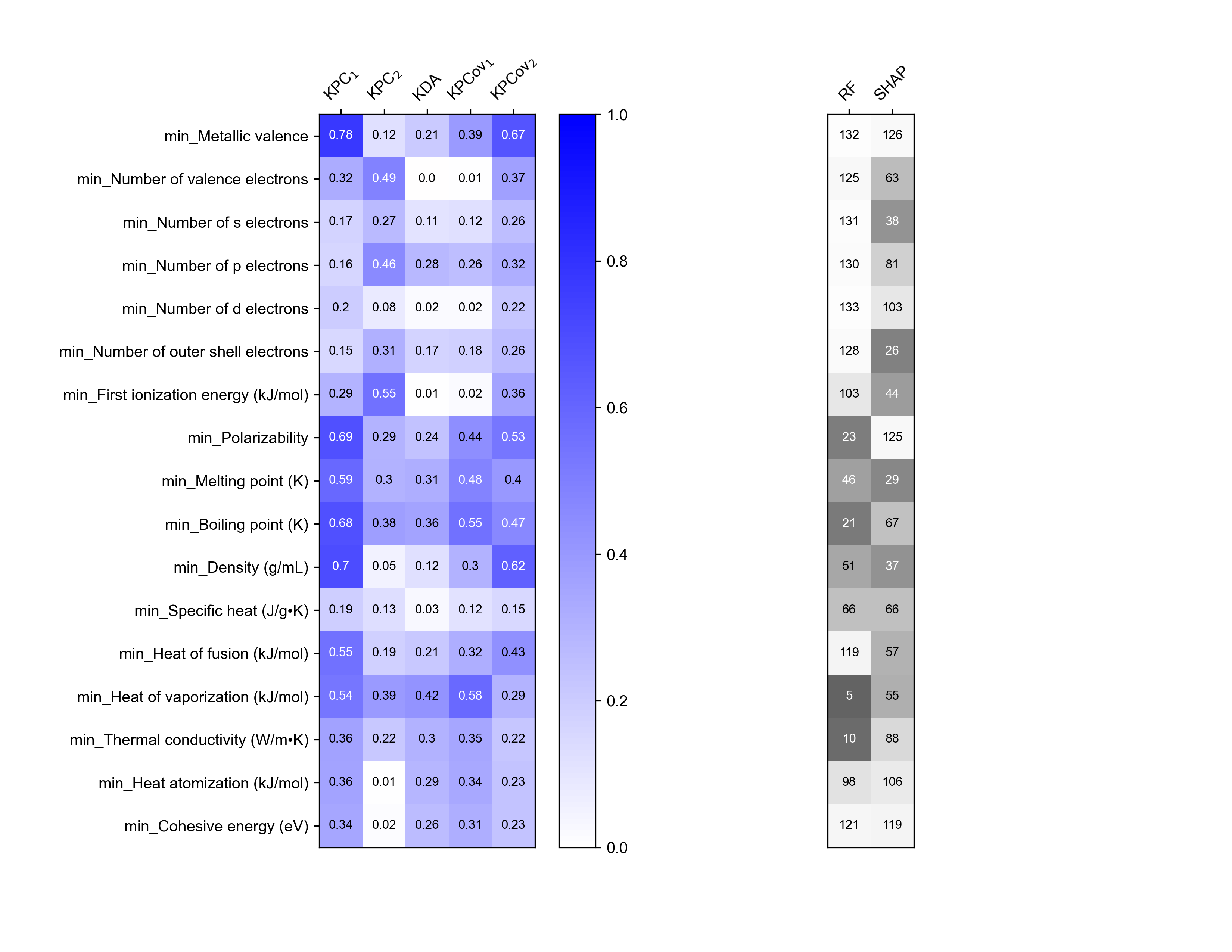}
\end{figure}

\bibliography{references}